\documentclass{article}
\usepackage{adjustbox}

\PassOptionsToPackage{numbers, sort&compress}{natbib}

\usepackage[preprint]{neurips_2026}

\usepackage[utf8]{inputenc} 
\usepackage[T1]{fontenc}    
\usepackage{hyperref}       
\hypersetup{colorlinks=true, allcolors=blue}
\usepackage{url}            
\usepackage{booktabs}       
\usepackage{amsfonts}       
\usepackage{nicefrac}       
\usepackage{microtype}      
\usepackage{xcolor}         
\usepackage{amsmath}
\usepackage{multirow}
\usepackage{makecell}
\usepackage[table]{xcolor}
\usepackage{graphicx}
\usepackage{algorithm}
\usepackage{algorithmic}
\usepackage[most]{tcolorbox}
\usepackage{wrapfig}
\usepackage{enumitem}
\usepackage{booktabs}
\usepackage{tabularx,graphicx,float}
\usepackage[most]{tcolorbox}

\tcbset{
  sliccase/.style={
    enhanced,
    breakable,
    width=\linewidth,
    boxrule=0.5pt,
    arc=1mm,
    boxsep=0pt,
    left=2.5mm,
    right=2.5mm,
    top=1.5mm,
    bottom=1.5mm,
    toptitle=1.2mm,
    bottomtitle=1.2mm,
    fonttitle=\small\bfseries,
    coltitle=black,
    fontupper=\small,
    before upper={\raggedright\setlength{\parindent}{0pt}},
    before skip=8pt,
    after skip=8pt
  }
}

\newtcolorbox{taskbox}[1]{
  sliccase,unbreakable,title={#1},
  colback=blue!2,colframe=blue!35!black,
  colbacktitle=blue!10
}
\newtcolorbox{obsbox}[1]{
  sliccase,unbreakable,title={#1},fontupper=\fontsize{8}{9}\selectfont,
  colback=black!2,colframe=black!40,
  colbacktitle=black!12
}
\newtcolorbox{reasonbox}[1]{
  sliccase,unbreakable,title={#1},fontupper=\fontsize{8}{9}\selectfont,
  colback=orange!3,colframe=orange!45!black,
  colbacktitle=orange!12
}
\newtcolorbox{resultbox}[1]{
  sliccase,unbreakable,title={#1},
  colback=green!2,colframe=green!35!black,
  colbacktitle=green!10
}

\title{SLIC: Reinforcement Fine-Tuning Small LMs for Multi-Turn Analog Circuit Optimization}

\author{%
\textbf{Yan Xu\textsuperscript{1},\hspace{0.8em}
Zhengqi Gao\textsuperscript{1},\hspace{0.8em}
Ching-Yun Ko\textsuperscript{3},\hspace{0.8em}
Zhang-Wei Hong\textsuperscript{3},\hspace{0.8em}
Wenjie Lu\textsuperscript{2}}\\[0.2em]
\textbf{Tao Yu\textsuperscript{2},\hspace{0.8em}
Xin Zhang\textsuperscript{3},\hspace{0.8em}
Duane S. Boning\textsuperscript{1},\hspace{0.8em}
Ruonan Han\textsuperscript{1}}\\[0.6em]
{
\textsuperscript{1}MIT
\quad
\textsuperscript{2}Analog Devices, Inc.
\quad
\textsuperscript{3}IBM Research
}
}

\begin{document}

\maketitle

\begin{abstract}
Analog integrated circuit (IC) design is a multi-turn, simulator-grounded engineering task that demands iterative reasoning over coupled performance trade-offs, yet proprietary process design kits (PDKs) make hosted large language model workflows difficult to deploy in practical design settings. We present \textbf{SLIC} -- \textit{\textbf{\underline{S}}mall
\textbf{\underline{L}}M \textbf{\underline{I}}terating on
\textbf{\underline{C}}ircuits}~-- a reinforcement fine-tuning framework that trains a 4B-scale, locally deployable language model for closed-loop analog circuit optimization. SLIC pairs structured circuit representations with discretized relative-update actions, 
domain- and success-gated multi-objective rewards, and 
Reward-Guided Rollout Steering (RGRS), which turns per-step simulator feedback into signals  for both policy learning and online rollout construction in this sparse, highly constrained domain. We further release an open, goal-conditioned benchmark of 
41 amplifier topologies with standardized protocols for 
in-distribution validation, held-out topology evaluation, and 
cross-technology-node transfer. Trained solely with open source 65\,nm CMOS predictive technology models (PTM), SLIC-\textsc{Random} achieves 87.1\% in-distribution four-specification Pass@1  within eight turns, surpassing the 763B DeepSeek-V4.1-Flash \textsc{Think} by 
48.5 percentage points. 
Beyond the training distribution, SLIC-\textsc{Softmax} reaches 72.6\% Pass@1 on held-out topologies, and the trained policies consistently transfer across five unseen technology nodes without additional fine-tuning. This transfer extends to a commercial TSMC 65~nm CMOS
PDK, where SLIC-\textsc{Random} achieves 49.9\%
held-out-topology Pass@1, demonstrating practical
deployment through local inference without transmitting
proprietary data to external LLM services.
Performance further scales with test-time simulator
interactions, reaching up to 83.2\% held-out-topology
Pass@1 at a 32-turn budget, suggesting that SLIC enables
the small policy to convert additional simulator calls
into improved circuit designs. Project page: \url{https://hangroup.mit.edu/slic/}.
\end{abstract}

\section{Introduction}
\label{sec:intro}
\begin{figure*}[t]
  \centering
  \includegraphics[width=\textwidth]{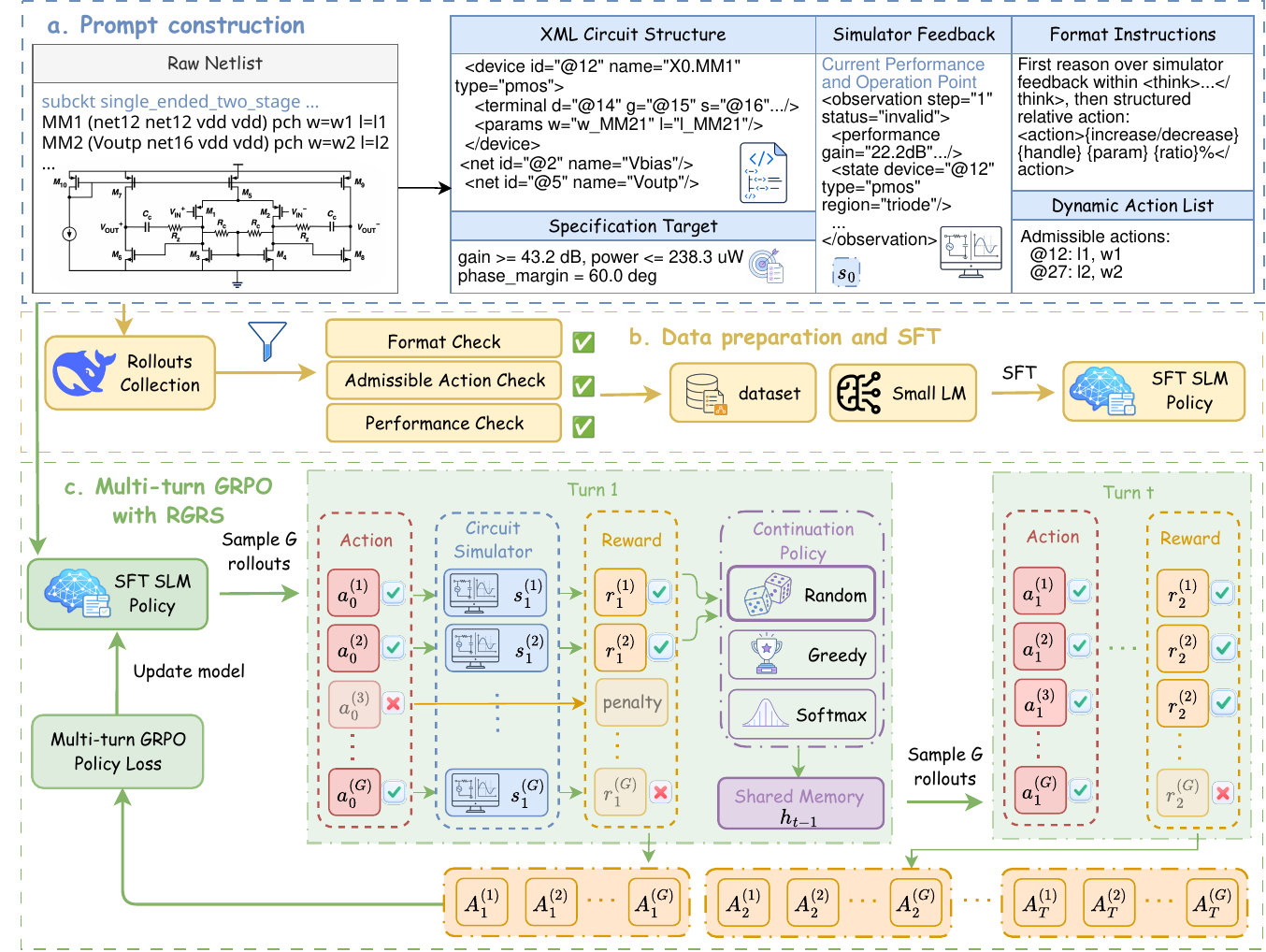}
  \vspace{-1.2em}
  \caption{
  Overview of \textbf{SLIC}. Given a goal-conditioned analog circuit optimization task, SLIC builds a structured prompt from the circuit representation and target specifications, then trains a language-model agent through SFT and simulator-interactive RL. 
    The upper path shows SFT data construction from filtered teacher rollouts. 
    The lower loop shows multi-turn GRPO with Reward-Guided Rollout Steering (RGRS), where multiple candidate responses propose
relative parameter updates that are evaluated through
simulation, one eligible continuation is selected to advance the
trajectory, and all candidates contribute to
group-relative policy updates. Subscripts denote interaction steps, and superscripts
index candidates within each group.
  }
  \label{fig:slic_overview}
  \vspace{-1.3em}
\end{figure*}

Analog integrated circuits (ICs)~\citep{carusone2011analog,razavi2017analog} sit at the interface between the physical world and modern computation. 
They perform high-fidelity conditioning and conversion of sensor signals~\citep{rao2025comprehensive}, 
enable efficient power regulation in advanced integrated systems~\citep{jain2024voltage}, 
support high-speed data conversion~\citep{bheemisetti2025saradc}, 
and form the high-bandwidth, low-power I/O fabric of AI accelerators and wireless infrastructure~\citep{mondal2024vcsel,smith2024mi300,wang2025futureg}. Despite this importance, analog IC design remains highly manual, expert-driven, and tightly coupled to process technology, making design productivity difficult to scale across circuits, specifications, and technology nodes. As artificial intelligence (AI) advances rapidly, this long-standing scalability bottleneck has renewed interest in AI-driven analog circuit design automation.

However, translating this opportunity into practical analog IC design frameworks remains challenging, as several key issues must be addressed.
First, any practical framework must \textbf{contend with severe data scarcity and semiconductor technology fragmentation.} Circuit performance and design decisions depend strongly on the underlying fabrication process, yet publicly available analog circuit designs are sparse due to proprietary protection and foundry confidentiality constraints. Many existing optimization~\citep{lyu2018batch,eriksson2019turbo,zhengqi2019bo} and learning-based approaches~\citep{settaluri2020autockt,wang2020gcnrl, cao2022domain, dong2023cktgnn, shahane2023graph} remain specialized to particular circuit families or rely primarily on numerical search, limiting their ability to transfer across design settings. Therefore, a \textbf{reasoning-driven approach trained on synthetic, simulator-grounded data} is preferable: a model learns to navigate the design space rather than to memorize specific solutions, which in turn enables \textbf{generalization across circuit topologies and technology nodes}. When implemented using LLMs, \textbf{human-interpretable design behavior} is enabled, where proposed decisions and updates throughout the design process can be understood, inspected, and refined rather than treated as opaque black-box outputs. 
Second, \textbf{local deployability} is essential: exposing foundry process design kits (PDKs), device models, or proprietary circuit inventions to external LLM APIs is prohibited in most commercial and national-security-related settings~\citep{lai2024analogcoder, liu2025eesizer,kochar2025ledro}. Repeated reliance on external LLM APIs in agentic design workflows also incurs substantial token consumption and operational costs~\citep{bai2026agentspend,wadlom2026efficient}. Third, the framework should support \textbf{scalability and parallelism}, enabling multiple specialized agents to operate concurrently on different circuit sub-modules (e.g. amplifiers, mixers and comparators) and thereby improve system-level design efficiency. It should also remain composable within a unified framework that can scale to larger systems with additional specialized agents, so that updating one agent when new circuit-domain-specific data or methodologies become available does not disturb the rest of the framework.

In this paper, we introduce \textbf{SLIC} (Small LM Iterating on Circuits, as shown in Figure~\ref{fig:slic_overview}), a lightweight, circuit-specific and reasoning-driven small language model (SLM) training framework that addresses the local deployability and scalability challenges above by performing multi-turn reinforcement learning with verifiable rewards (RLVR). RLVR has recently driven substantial progress in single-turn reasoning tasks such as mathematics and coding~\citep{shao2024deepseekmath,deepseekr1,jaech2024openai_o1,kazemnejad2024vineppo,cui2025process,yu2025dapo,zha2025rl,openai2024learning,trung2024reft,chen2024step,uesato2022solving,wei2025swerl}, and has been extended to multi-turn agentic settings~\citep{feng2025gigpo,li2025turnppo} where models act over multiple steps and adapt to environmental feedback~\citep{yao2022webshop,shridhar2020alfworld}. Such a paradigm shift from isolated answer generation toward stateful decision-making is a natural fit for analog circuit design: it is inherently multi-turn, requires reasoning over coupled trade-offs among performance objectives, and admits quantitatively verifiable feedback from circuit simulators at every step. 
Training small LMs under this paradigm reflects a practical trade-off between reasoning capability and parallel deployability. 
We focus on 4B-scale models as a balanced operating point: large enough to support structured reasoning and action generation, yet small enough to enable multiple specialized agents to run concurrently in local design environments. Our contributions are threefold:

\begin{enumerate}[
  labelindent=0pt,
  leftmargin=*,
  align=left,
  labelsep=0.5em
]
\item \textbf{A reasoning-driven, locally deployable framework for 
    analog design optimization.} We train SLIC on simulator-grounded synthetic data
generated using public predictive technology models (PTM), addressing data
scarcity without requiring proprietary PDK data.
Trained solely using PTM 65~nm CMOS, the 4B SLIC-\textsc{Softmax}
policy maintains 61.9--71.0\% Pass@1 on held-out
topologies across five unseen technology nodes without
additional fine-tuning, compared with 72.6\% at the
training node.
Transfer to a commercial TSMC 65~nm CMOS PDK further
demonstrates practical deployment through local inference
and simulation without exposing proprietary process
models or circuit data to external LLM services.

    \item \textbf{Domain adaptations for simulator-interactive
design optimization.}
We combine domain- and success-gated rewards with
Reward-Guided Rollout Steering (RGRS) to learn from
intermediate simulator feedback.
The rewards prioritize functional designs and target
satisfaction, while RGRS guides rollout continuation
and supports comparisons among candidate actions
from the same interaction history.
In our main setting, the 4B SLIC-\textsc{Random} policy
achieves 87.1\% in-distribution four-specification
Pass@1 within eight turns, outperforming the 763B
DeepSeek-V4.1-Flash \textsc{Think} baseline by
48.5 percentage points.
    
    \item \textbf{Open simulator-interactive benchmark.} We release an 
    open, goal-conditioned benchmark of 41 amplifier topologies paired 
    with target specifications and initial states, with standardized 
    protocols for in-distribution validation, held-out topology 
    evaluation, and cross-technology-node transfer (22--130\,nm CMOS), 
    providing a reproducible testbed for closed-loop, multi-turn analog 
    circuit design optimization.
\end{enumerate}

\section{Background and Problem Formulation}
\label{sec:background}
\subsection{Multi-turn RL for Analog Circuit Optimization}

We formulate analog circuit optimization as a multi-turn RLVR problem. 
At turn $t$, let $c_t \in \mathcal{C}$ denote the textual circuit 
representation and $f_t \in \mathcal{F}$ the simulator feedback, 
comprising measured performance metrics (e.g., gain, power) and device-level operating states (e.g., 
whether each transistor is in saturation or triode). The stepwise 
state is $s_t = (c_t, f_t)$. Let $h_t$ contain the task specification $q$, the current
state $s_t$, and the retained interaction history.
The policy generates a variable-length sequence of
circuit edit actions $a_t \sim \pi_\theta(\cdot \mid h_t)$
corresponding to a structured parameter update. The 
simulator-interactive environment $\mathcal{M}$ applies the update 
and returns $s_{t+1} = \mathcal{M}(s_t, a_t)$ together with a 
verifiable reward $r_t = \mathcal{R}(a_t, s_{t+1})$. An episode 
terminates when all target specifications are satisfied or when 
the maximum horizon is reached.

\subsection{Multi-turn GRPO}
Given a task instance $(q,s_0)$, trajectory-level GRPO~\citep{shao2024deepseekmath} samples a group of trajectories $\{\tau_i\}_{i=1}^G$ from the behavior policy $\pi_{\theta_{\mathrm{old}}}$, where each trajectory $\tau_i=(s_{0,i},a_{0,i},r_{0,i},\dots,s_{T_i,i})$ is generated through simulator-interactive optimization. Let $o_i=(o_{i,1},\dots,o_{i,|o_i|})$ denote the model output sequence that induces trajectory $\tau_i$ through interaction with the environment. Each sampled trajectory $\tau_i$ receives a scalar
return $R_i=R(\tau_i)$, defined by its terminal reward
$R_i=r_{T_i-1,i}$ or by aggregating step rewards as
$R_i=\sum_{t=0}^{T_i-1}r_{t,i}$.
GRPO normalizes these returns across the sampled
trajectories:
\begin{equation}
\hat A_i
=
\frac{R_i-\mu_G}{\sigma_G+\epsilon},
\qquad
\mu_G=\frac{1}{G}\sum_{i'=1}^{G}R_{i'},
\qquad
\sigma_G=
\operatorname{std}\!\left(\{R_{i'}\}_{i'=1}^{G}\right),
\label{eq:group_adv}
\end{equation}
where $\epsilon>0$ ensures numerical stability.
The resulting advantage is shared across all turns
and generated response tokens $\ell$ in the trajectory,
so $\hat A_{i,\ell}=\hat A_i$.
Turn-level variants retain the clipped policy objective
below but construct a separate advantage for each turn.

The policy is then updated with a PPO-style clipped objective~\citep{schulman2017ppo},
\begin{equation}
\begin{aligned}
\mathcal{L}_{\mathrm{GRPO}}(\theta)
=
&\mathbb{E}\!\left[
\frac{1}{G}\sum_{i=1}^{G}\frac{1}{|o_i|}\sum_{\ell=1}^{|o_i|}
\min\!\Big(
\rho_{i,\ell}(\theta)\hat A_{i,\ell},\;
\mathrm{clip}(\rho_{i,\ell}(\theta),1-\epsilon_{\mathrm{low}},1+\epsilon_{\mathrm{high}})\hat A_{i,\ell}
\Big)
\right] \\
&\quad - \beta D_{\mathrm{KL}}(\pi_\theta \,\|\, \pi_{\mathrm{ref}}),
\end{aligned}
\label{eq:multi_turn_grpo}
\end{equation}
where $\rho_{i,\ell}(\theta)=\pi_\theta(o_{i,\ell}\mid\cdot)/\pi_{\theta_{\mathrm{old}}}(o_{i,\ell}\mid\cdot)$ is the token-level importance ratio, with both policies conditioned on the same context and preceding tokens. We set $\beta=0$ (Table~\ref{tab:rl_config}). Following the Clip-Higher strategy in DAPO~\citep{yu2025dapo}, we use an asymmetric clipping range with $\epsilon_{\mathrm{high}} > \epsilon_{\mathrm{low}}$, mildly relaxing the upper policy-ratio bound for positive-advantage tokens to preserve exploration during multi-turn training.

\subsection{Challenges of Off-the-Shelf Multi-Turn GRPO
for Analog Circuit Design Optimization}
\label{sec:challenges}

Although multi-turn GRPO provides a natural
policy-optimization backbone for simulator-interactive
circuit design optimization, applying it directly
presents several challenges.
First, analog circuit optimization involves tightly
coupled performance requirements within a narrow
feasible design space.
Assessing progress requires considering both circuit
validity and improvements toward satisfying all
target specifications.
Second, vanilla trajectory-level credit assignment
can be ambiguous in multi-turn optimization.
Since the final trajectory return is attached to
multiple turns, invalid or unhelpful turns may inherit
high rewards from trajectories that are later rescued,
while locally beneficial turns may receive low scores
when later turns fail.
This motivates denser intermediate rewards that provide
more direct feedback on individual updates.
Finally, sampled groups can lack reward contrast,
including when all candidates fail or all succeed.
If their rewards are identical, group-relative
advantages vanish.
Maintaining informative reward differences within
each group is therefore important for learning
efficiently from sampled simulator feedback.

\section{SLIC: Domain Adaptations for Simulator-Interactive Circuit Optimization} 
\label{headings}

\subsection{SLIC Framework}
\label{sec:pipeline}
Figure~\ref{fig:slic_overview} summarizes the SLIC pipeline. We 
collect training trajectories by running a strong prompted model 
(DeepSeek-V4.1-Flash) on simulator-grounded multi-turn optimization without 
specifying explicit goals, then relabel each trajectory with its 
best-observed performance tuple as the goal-conditioned training 
target, following the hindsight relabeling principle of 
HER~\citep{andrychowicz2017her}. This bypasses the lack of labeled 
(circuit, target-spec) pairs and yields a fully simulator-grounded 
synthetic training set. The dataset is 
filtered to retain valid and performance-improving turns for 
supervised fine-tuning (SFT), which initializes the policy with 
basic reasoning and action-generation behavior 
(Figure~\ref{fig:slic_overview}b). The SFT checkpoint is then 
further trained with multi-turn GRPO using the domain-gated rewards 
of Section~\ref{sec:reward} and the Reward-Guided Rollout Steering 
procedure of Section~\ref{sec:branch_continuation} 
(Figure~\ref{fig:slic_overview}c).

Each prompt contains the XML-structured circuit state, target specifications, simulator feedback, and a dynamically generated admissible action list (Figure~\ref{fig:slic_overview}a). 
The policy outputs reasoning and interpretable discretized relative-update actions rather than absolute parameter values. Full representation details, action grammar, and validity checks are provided in Appendix~\ref{app:representation}.

\subsection{Domain Reward Shaping}
\label{sec:reward}

Analog circuit optimization requires satisfying coupled
performance targets under physically meaningful operating
conditions.
Our reward design prioritizes valid amplification and joint
specification satisfaction, while rewarding additional
performance headroom among successful designs.

\paragraph{Domain-validity gates.}
Directly aggregating performance metrics can favor
nonfunctional designs, such as circuits that consume
little power because they fail to provide valid amplification.
We therefore apply gates that reflect the physical
prerequisites of amplifier operation before aggregating
per-specification rewards.
Non-positive gain suppresses power, bandwidth, and
phase-margin credit.
An in-band flatness gate additionally suppresses gain,
bandwidth, and phase-margin credit when peaking exceeds
6~dB, preventing distorted frequency responses from
receiving misleading performance credit.
The same gates apply when checking specification
satisfaction, with flatness used only to assess validity.
Full definitions are provided in
Appendix~\ref{app:reward_details}.

\paragraph{Success-conditional slack bonus.}
To prioritize target satisfaction, we keep unsuccessful
scores strictly below one and assign successful
scores in $[1,1.5]$, with a bonus based on mean normalized
specification slack.
A constant success reward would leave all-success groups
without a performance learning signal.
DAPO~\citep{yu2025dapo} addresses this through dynamic
sampling, filtering out groups with uniform correctness
outcomes.
To obtain richer feedback from successful terminal
candidates, we use the slack bonus to preserve differences
in their measured specification headroom, allowing
all-success groups to remain informative when their
slack differs.
The bonus also encourages additional headroom, providing
margin against performance degradation and improving design robustness and policy generalization.
The formulation and its tolerance alignment are detailed
in Appendix~\ref{app:reward_details}.

For policy updates, we normalize validity and performance
feedback separately before combining their advantages,
following the reward-decoupled normalization principle
of GDPO~\citep{liu2026gdpo}.
The formulation is provided in
Appendix~\ref{app:separate_normalization}.

\subsection{Reward-Guided Rollout Steering}
\label{sec:branch_continuation}

Motivated by the credit-assignment challenges discussed
in Section~\ref{sec:challenges}, we introduce
\textbf{Reward-Guided Rollout Steering (RGRS)},
a search-augmented procedure for online rollout
construction guided by intermediate simulator feedback.
At each turn, RGRS samples $G$ candidate responses
from the same interaction history, each proposing
an alternative set of parameter edits, and evaluates
the resulting circuits through simulation.
This procedure serves two roles.
For policy learning, all sampled candidates remain
available for group-relative updates, providing
same-context comparisons for stepwise advantage estimation.
For trajectory construction, RGRS selects one eligible
successor as the parent state for the next turn,
advancing the shared history.
Continuation selection is restricted to valid candidates
whenever any are available.

This design is motivated by the view of search as a
policy-improvement operator.
Best-of-$G$ selection induces a reward-biased distribution over
outputs, and prior work has analyzed the distributional shift and
KL trade-offs of such best-of-$N$
policies~\citep{beirami2024theoretical}.
More broadly, expert-iteration and AlphaZero-style methods use
search to produce improved behavior that is later internalized by
a learned policy~\citep{danihelka2022policy}.
A related principle appears in automated discovery:
TTT-Discover~\citep{yuksekgonul2026learning} selects promising parent states from previously
evaluated candidates while retaining exploration through its
reuse procedure.
In our setting, RGRS applies candidate selection locally at each turn.
We study three continuation policies to examine the trade-off
between performance preference and exploration:
\textsc{Random} samples uniformly from eligible candidates,
\textsc{Greedy} selects a highest-scoring candidate,
and \textsc{Softmax} favors higher-scoring candidates while retaining
nonzero selection probability for other eligible candidates.
The full procedure is provided in
Algorithm~\ref{alg:reward_guided_continuation}
in Appendix~\ref{app:rgrs_algo}.

\section{Experiments}
\label{sec:exp}
\subsection{Setup}
\label{sec:setup}
\paragraph{Dataset.}
We construct a benchmark of 41 amplifier topologies,
using 35 for training and in-distribution (ID) evaluation
and reserving six held-out topologies for out-of-distribution (OOD) evaluation. The dataset is further divided into Simple and Complex topology categories, each paired with multiple target goals and initial states. As shown in Table~\ref{tab:standard_hard_stats}, the benchmark spans a wide range of topology complexity, tunable parameter counts, and valid/invalid initial-state ratios, providing a challenging testbed for multi-turn analog circuit optimization. The main experiments extend the original evaluation with
a newer backbone, a larger benchmark, and an additional
bandwidth objective.
Training at 65\,nm enables transfer evaluation toward
both more advanced (22–45\,nm) and more mature (90–130\,nm) technology nodes.
Appendix~\ref{app:original_evaluation} reports this earlier
configuration, which uses the Apache-2.0-licensed Qwen2.5-7B-Instruct~\citep{qwen2p5_7b_instruct, qwen2024qwen25},
22\,nm training, and three specifications (excluding bandwidth).

\paragraph{Simulation environment.}
Training tasks are simulated with ngspice~\citep{ngspice}
using the 65~nm bulk CMOS Predictive Technology Model
(PTM)~\citep{zhao2006ptm, ptm_website}.
In this setting, transistor sizing is constrained to
channel length \(L \in [45\,\mathrm{nm}, 100\,\mu\mathrm{m}]\)
and width \(W \in [45\,\mathrm{nm}, 500\,\mu\mathrm{m}]\).
For cross-technology evaluation, each node uses its
corresponding PTM models~\citep{ptm_website} and
node-specific sizing bounds, as summarized in
Table~\ref{tab:tech_node_ranges}.
We use Cadence Spectre~\citep{cadenceSpectre} for
evaluation with a commercial TSMC 65~nm CMOS foundry PDK.

\paragraph{Model training configuration.}
We initialize SFT from Apache-2.0-licensed
Qwen3.5-4B~\citep{qwen3.5,qwen35_4b_model},
choosing a 4B-scale model as a practical operating point
for locally deployable specialized analog design agents
under proprietary PDK constraints.
RL is initialized from the resulting SFT checkpoint
and optimized with GRPO using RGRS.
Training uses an XML circuit representation and, unless another
horizon is specified, a maximum
rollout horizon of eight turns, taking approximately
39 hours on 2$\times$B200 GPUs.
Tables~\ref{tab:sft_config} and~\ref{tab:rl_config}
summarize the SFT and RL configurations, respectively. We also train a variant of SLIC-\textsc{Random} with
dynamic filtering inspired by DAPO's dynamic
sampling~\citep{yu2025dapo}, which skips candidate groups with no group relative policy gradient signal. It is labeled ``+ filter'' in the figures and
``+ dynamic filtering'' in the tables. To assess applicability beyond Qwen3.5-4B, we also apply
the same SFT-to-RL pipeline to the Apache-2.0-licensed
Qwen3.5-9B~\citep{qwen3.5,qwen35_9b_model}
and Qwen3-8B~\citep{qwen3technicalreport,qwen3_8b_model}
using SLIC-\textsc{Random}.
Results are reported in Table~\ref{tab:backbone_generalization}.

\paragraph{Baselines.}
We compare our method with several complementary baselines. We include Bayesian Optimization with TuRBO~\citep{eriksson2019turbo} as a representative black-box numerical optimizer, the Qwen3.5-4B model to evaluate the zero-shot capability of an untuned locally deployable LLM, and the supervised fine-tuned (SFT) model as an initialization without reinforcement learning. We also evaluate DeepSeek-V4.1-Flash~\citep{deepseek2026v41flash,deepseek_v41_flash_model}, a recent large open-weight model, in \textsc{Think} mode as a strong prompt-based reasoning baseline. To compare SLIC with a strong closed-source model,
we evaluate GPT-6-Astra~\citep{openai2026gpt6astra} on the same tasks, with one rollout per task
and up to eight interaction turns. See Appendix~\ref{app:hosted_model_eval} for evaluation notes
and API costs for both models. We also include two GRPO baselines.
Single-turn GRPO uses SLIC's reward design,
with the effect of training horizon analyzed in
Section~\ref{sec:backbone_generalization}.
For multi-turn training, trajectory-level GRPO with
only terminal supervision was unstable in our experiments.
We therefore evaluate the branching-rollout formulation of multi-turn GRPO (MT-GRPO)~\citep{zeng2025mtgrpo}, referred to here as
Tree MT-GRPO, which combines intermediate and
terminal feedback, with SLIC's success-conditional slack bonus in the outcome reward.
Its budget of candidate evaluations is approximately
matched to SLIC-\textsc{Random}.
Implementation details are provided in
Appendix~\ref{app:tree_mt_grpo}.

\paragraph{Metrics.}
We evaluate four specifications: gain, power consumption,
$-3\,\mathrm{dB}$ bandwidth, and phase margin.
A rollout succeeds if it reaches a valid design satisfying
all four target specifications within the prescribed 5\%
tolerance under the domain-validity gates.
At a turn budget $T$, Pass@$k$ measures the probability
that at least one of $k$ sampled rollouts succeeds within
$T$ interaction turns, averaged over evaluation tasks.
We primarily report cumulative four-specification Pass@1
at Turns~4 and~8.
Unless otherwise noted, we report means $\pm$ standard
deviations over three evaluation runs.
\subsection{Evaluation Results}

\paragraph{In-distribution evaluation.}
Table~\ref{tab:val_turns} shows that all 4B SLIC variants
substantially outperform SFT and DeepSeek-V4.1-Flash
\textsc{Think} at both evaluation horizons.
At four turns, SLIC-\textsc{Random} achieves 74.9\%
Pass@1, exceeding GPT-6-Astra's 67.1\%.
At eight turns, Astra achieves higher success at 90.6\%,
while the locally deployable 4B SLIC-\textsc{Random}
policy reaches 87.1\%, a gap of 3.5 percentage points.
Among SLIC variants, \textsc{Softmax} performs best on simple topologies,
while \textsc{Random} leads on complex topologies and overall.
Although DeepSeek-V4.1-Flash also generated the SFT data, RL lifts
the student from below its teacher (26.5\% vs.\ 38.6\%) to well
above it (87.1\%).

\begin{table*}[t]
\centering
\small
\setlength{\tabcolsep}{3pt}
\caption{
In-distribution evaluation at the PTM 65\,nm
technology node over 840 tasks with breakdowns by topology complexity. GPT-6-Astra is evaluated once on the full evaluation set.
Bayesian Optimization uses 96 and 192 simulator calls per topology to match SLIC's total simulation budget.
\textbf{Bold} and \underline{underline} denote the best and
second-best mean results in each column respectively.
}
\label{tab:val_turns}
\vspace{2pt}
\begin{adjustbox}{max width=\textwidth}
\begin{tabular}{@{}lcccccc@{}}
\toprule
& \multicolumn{3}{c}{\textbf{4-spec Pass@1 (\%) @Turn 4}}
& \multicolumn{3}{c}{\textbf{4-spec Pass@1 (\%) @Turn 8}} \\
\cmidrule(lr){2-4}
\cmidrule(lr){5-7}
\textbf{Method}
  & \textbf{Simple} & \textbf{Complex} & \textbf{Total}
  & \textbf{Simple} & \textbf{Complex} & \textbf{Total} \\
\midrule
Bayesian Optimization
  & $54.3 \pm 7.2$
  & $34.2 \pm 1.9$
  & $43.4 \pm 3.9 $
  & $70.5 \pm 7.5$
  & $54.1 \pm 5.3$
  & $61.6 \pm 6.1$ \\
\midrule
GPT-6-Astra
& 77.6 & 58.3 & 67.1
& $96.1$ & $\mathbf{86.0}$ & $\mathbf{90.6}$ \\
DeepSeek-V4.1-Flash \textsc{Think} (763B)
& $33.1 \pm 0.0$ & $11.1 \pm 1.8$ & $21.2 \pm 1.0$
& $56.0 \pm 1.6$ & $24.0 \pm 1.1$ & $38.6 \pm 1.3$ \\
\midrule
Qwen3.5-4B
  & $2.4 \pm 0.6$
  & $1.0 \pm 0.3$
  & $1.7 \pm 0.2$
  & $3.2 \pm 1.1$
  & $1.5 \pm 0.4$
  & $2.3 \pm 0.5$ \\
\quad SFT
  & $22.4 \pm 1.2$
  & $6.1 \pm 0.1$
  & $13.5 \pm 0.6$
  & $41.0 \pm 1.8$
  & $14.4 \pm 0.7$
  & $26.5 \pm 0.7$ \\
\quad Single-turn GRPO
  & $66.4 \pm 2.0$
  & $41.7 \pm 0.8$
  & $53.0 \pm 1.0$
  & $74.3 \pm 0.8$
  & $51.0 \pm 1.9$
  & $61.6 \pm 1.4$ \\
\quad Tree MT-GRPO
  & $82.7 \pm 1.6$ & $49.6 \pm 2.5$ & $64.7 \pm 0.9$
  & $\mathbf{96.3 \pm 1.1}$ & $68.4 \pm 1.4$ & $81.2 \pm 1.3$ \\
\midrule
\multicolumn{7}{@{}l}{\textit{Proposed (Qwen3.5-4B)}} \\
\quad SLIC-\textsc{Greedy}
  & $72.4 \pm 1.3$
  & $36.6 \pm 1.5$
  & $53.0 \pm 1.2$
  & $84.1 \pm 0.9$
  & $51.6 \pm 1.1$
  & $66.5 \pm 0.8$ \\
\rowcolor{gray!12}
\quad \textbf{SLIC-\textsc{Random}}
  & $\underline{86.8 \pm 0.8}$
  & $\mathbf{64.8 \pm 0.7}$
  & $\mathbf{74.9 \pm 0.0}$
  & $92.7 \pm 1.2$
  & $\underline{82.5 \pm 1.0}$
  & $\underline{87.1 \pm 0.7}$ \\
\qquad w/o domain-validity gates
& $82.2 \pm 0.8$
& $57.2 \pm 0.1$
& $68.6 \pm 0.4$
& $89.1 \pm 0.5$
& $77.6 \pm 2.3$
& $82.8 \pm 1.2$ \\
\quad SLIC-\textsc{Softmax}
  & $\mathbf{87.7 \pm 0.3}$
  & $\underline{59.8 \pm 1.5}$
  & $\underline{72.5 \pm 0.7}$
  & $\underline{93.9 \pm 0.8}$
  & $76.2 \pm 1.5$
  & $84.3 \pm 1.0$ \\

\bottomrule
\end{tabular}
\end{adjustbox}
\vspace{-0.8em}
\end{table*}

\paragraph{Generalization to held-out topologies.}
Table~\ref{tab:test_topo_turns} shows that SLIC generalizes
to held-out topologies.
At eight turns, SLIC-\textsc{Softmax} achieves
the highest Pass@1 among SLIC variants, exceeding DeepSeek by 45.9 percentage points.
GPT-6-Astra achieves 86.2\% on the same held-out tasks,
with its advantage concentrated in structurally
simple topologies, which can still require substantial
performance improvements to meet their targets.
Both methods achieve 82.5\% on complex topologies.
Softmax's advantage over Random is not statistically
significant (Appendix~\ref{app:statistical_analysis}),
although both outperform Greedy, suggesting a benefit
from preserving diversity during rollout construction for topology generalization. Rollout analysis associates Softmax with more consistent
recovery and planning after performance decreases, while Greedy
shows more hesitation and repeated reconsideration
despite longer responses (Appendix~\ref{app:training_dynamics}, Figure~\ref{fig:training_dynamics}). Required performance improvements also help explain
differences across topologies beyond circuit size
alone (Appendix~\ref{app:topology_factors}).

\paragraph{Comparison with Tree MT-GRPO.}
While Tree MT-GRPO leads SLIC-\textsc{Random} on simple ID topologies at Turn~8 
(96.3\% vs.\ 92.7\%), SLIC-\textsc{Random} achieves higher overall
Pass@1 on both ID tasks
(87.1\% vs.\ 81.2\%) and held-out topologies
(70.6\% vs.\ 58.3\%), with the largest
gaps on complex topologies
(Tables~\ref{tab:val_turns} and~\ref{tab:test_topo_turns}). In circuit optimization, each action jointly modifies
multiple parameters, making it important to associate
the resulting performance with the corresponding update.
This association is less direct in Tree MT-GRPO, where aggregating advantages across subsequent
turns mixes feedback on the current joint update with
the effects of later edits.
Moreover, Tree MT-GRPO exhibits a sharp increase in policy entropy during training.
Its branching structure also makes sampling cost grow
exponentially with the training horizon, limiting
the depth of training rollouts. SLIC therefore uses RGRS to compare candidates from the same
interaction history and learns from stepwise simulator
feedback.
This supports effective iterative circuit optimization
by keeping credit assignment focused on local refinement,
while advancing one selected successor allows larger
candidate groups and longer training rollouts with
linear sampling cost.

\begin{table*}[t]
\centering
\small
\setlength{\tabcolsep}{4pt}
\renewcommand{\arraystretch}{1.08}
\caption{
Out-of-distribution evaluation
at the PTM 65~nm technology node over 240 tasks, with breakdowns
by topology complexity (two simple and four complex). GPT-6-Astra is evaluated once on the full evaluation set.
Bayesian Optimization uses 96 and 192 simulator calls per topology to match SLIC's total simulation budget.
\textbf{Bold} and \underline{underline} denote the best and
second-best reported mean results in each column respectively.
}
\label{tab:test_topo_turns}
\vspace{2pt}
\begin{adjustbox}{max width=\textwidth}
\begin{tabular}{@{}lcccccc@{}}
\toprule
& \multicolumn{3}{c}{\textbf{4-spec Pass@1 (\%) @Turn 4}}
& \multicolumn{3}{c}{\textbf{4-spec Pass@1 (\%) @Turn 8}} \\
\cmidrule(lr){2-4}
\cmidrule(lr){5-7}
\textbf{Method}
  & \textbf{Simple}
  & \textbf{Complex}
  & \textbf{Total}
  & \textbf{Simple}
  & \textbf{Complex}
  & \textbf{Total} \\
\midrule

Bayesian Optimization
  & $2.5 \pm 0.0$
  & $36.7 \pm 1.0$
  & $25.3 \pm 0.6$
  & $3.3 \pm 1.4$
  & $46.0 \pm 1.4$
  & $31.8 \pm 0.5$ \\
\midrule

GPT-6-Astra
& $\mathbf{66.2}$ & 62.5 & $\mathbf{63.7}$
& $\mathbf{93.8}$ & $\mathbf{82.5}$ & $\mathbf{86.2}$ \\

DeepSeek-V4.1-Flash \textsc{Think} (763B)
& $10.8 \pm 3.8$ & $16.9 \pm 1.1$ & $14.9 \pm 0.6$
& $20.0 \pm 3.7$ & $30.0 \pm 2.3$ & $26.7 \pm 2.2$ \\
\midrule

Qwen3.5-4B
  & $3.3 \pm 0.7$
  & $3.8 \pm 0.6$
  & $3.6 \pm 0.5$
  & $3.3 \pm 0.7$
  & $5.0 \pm 1.1$
  & $4.4 \pm 0.9$ \\

\quad SFT
  & $3.3 \pm 3.1$
  & $14.8 \pm 1.3$
  & $11.0 \pm 0.2$
  & $10.4 \pm 1.4$
  & $24.6 \pm 1.3$
  & $19.9 \pm 1.3$ \\
\quad Single-turn GRPO
  & $31.2 \pm 2.5$
  & $41.2 \pm 2.9$
  & $37.9 \pm 2.7$
  & $49.2 \pm 0.7$
  & $46.9 \pm 1.7$
  & $47.6 \pm 0.9$ \\
\quad Tree MT-GRPO
  & $29.6 \pm 5.8$ & $52.3 \pm 1.6$ & $44.7 \pm 1.0$
  & $45.0 \pm 3.3$ & $65.0 \pm 4.1$ & $58.3 \pm 3.6$ \\
\midrule
\multicolumn{7}{@{}l}{\textit{Proposed (Qwen3.5-4B)}} \\

\quad SLIC-\textsc{Greedy}
  & $22.5 \pm 2.5$
  & $36.2 \pm 2.9$
  & $31.7 \pm 1.8$
  & $32.1 \pm 1.9$
  & $49.8 \pm 2.9$
  & $43.9 \pm 2.4$ \\

\quad SLIC-\textsc{Random}
  & $35.0 \pm 1.2$
  & $\mathbf{67.7 \pm 2.2}$
  & $56.8 \pm 1.3$
  & $46.7 \pm 1.4$
  & $\mathbf{82.5 \pm 2.3}$
  & $70.6 \pm 1.2$ \\
\qquad w/o domain-validity gates
& $31.2 \pm 4.5$
& $55.2 \pm 2.2$
& $47.2 \pm 2.3$
& $46.2 \pm 2.2$
& $\underline{72.3 \pm 1.9}$
& $63.6 \pm 0.9$ \\
\rowcolor{gray!12}
\quad \textbf{SLIC-\textsc{Softmax}}
  & $\underline{36.7 \pm 4.0}$
  & $\underline{67.1 \pm 3.0}$
  & $\underline{56.9 \pm 3.1}$
  & $\underline{52.9 \pm 4.7}$
  & $\mathbf{82.5 \pm 0.6}$
  & $\underline{72.6 \pm 2.0}$ \\
\bottomrule
\end{tabular}
\end{adjustbox}
\vspace{-0.2em}
\end{table*}

\paragraph{Technology node generalization.}

\begin{figure*}[t]
    \centering
\includegraphics[width=\textwidth]{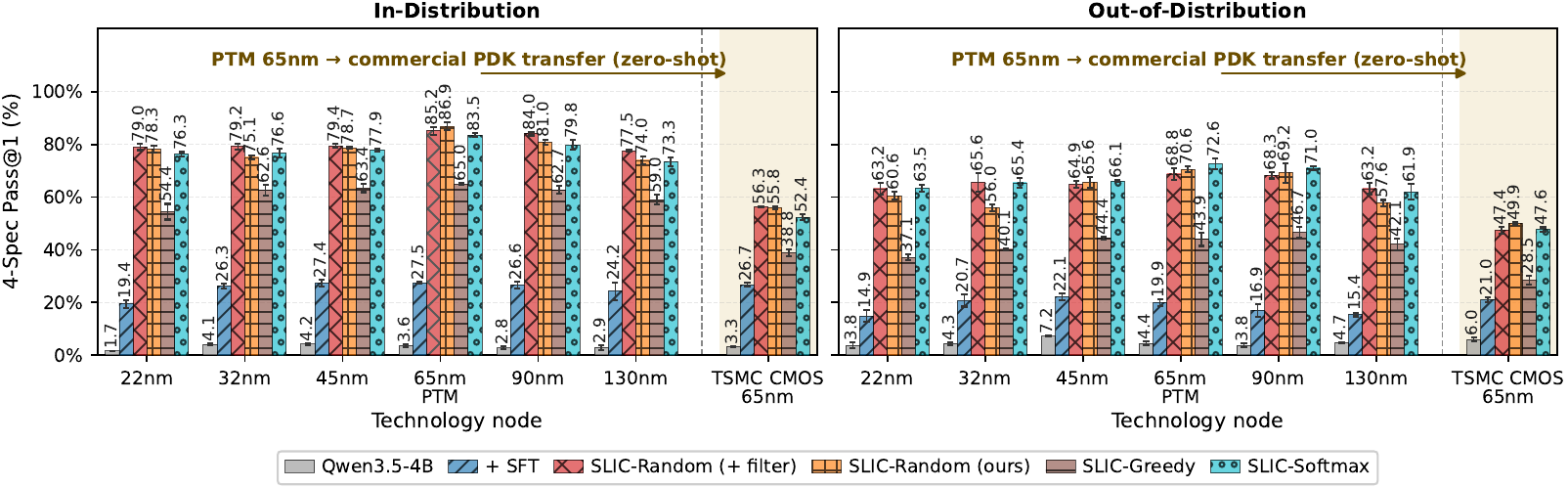}
    \vspace{-1.2em}
\caption{
Technology transfer under four-specification Pass@1
evaluation within eight turns.
Policies are trained on PTM 65~nm and evaluated across
PTM nodes from 22 to 130~nm and a TSMC CMOS 65~nm PDK
without additional fine-tuning.
Initial parameters are adjusted to the valid ranges
of each technology, while the PTM 65~nm target specifications
are reused.
Left: a 420-task subset of the full 840 in-distribution tasks.
Right: 240 held-out topology tasks.
}
    \label{fig:tech_node_sweep}
    \vspace{-1.0em}
\end{figure*}

Figure~\ref{fig:tech_node_sweep} shows that policies trained
at PTM 65~nm can transfer to both older and more advanced technology
nodes without additional fine-tuning.
Across the five unseen technology nodes, SLIC-\textsc{Softmax}
maintains 73.3--79.8\% Pass@1 on ID tasks and 61.9--71.0\%
on held-out topologies, compared with 83.5\% and 72.6\%,
respectively, at the training node. These results
suggest that SLIC internalizes simulator-grounded optimization
strategies that remain useful beyond the training node, supporting our hypothesis
that some learned strategies are agnostic to the technology node. Valid parameter
ranges for each node are reported in Table~\ref{tab:tech_node_ranges}.

To assess effectiveness in a practical circuit design
setting, we also evaluate SLIC on a commercial TSMC
65~nm CMOS foundry PDK without additional fine-tuning.
SLIC-\textsc{Random} achieves 55.8\% ID
and 49.9\% OOD Pass@1 within eight turns, down from 86.9\% and 70.6\%,
respectively, on the same tasks with PTM 65~nm. The foundry PDK captures
process-specific effects that PTM simplifies. Differences in device behavior can therefore shift the
trade-offs achievable with a given topology, making the four targets selected under PTM
harder to satisfy jointly. This mismatch may help explain the observed
drop in Pass@1 after transfer. Our task-level audit shows that SFT mainly succeeds when the initial
design needs only minor adjustments to meet the targets, which may
explain its similar Pass@1 across the two processes. On the foundry PDK, SLIC-\textsc{Random} still outperforms SFT
by about 29 percentage points on both splits.
These results demonstrate the practical value of local
policy inference and simulation, which keep proprietary
device models and circuit data in-house while enabling
automated circuit optimization in workflows where
PDK confidentiality prevents the use of hosted LLM services.

\subsection{Component Ablations}
\label{sec:component_ablations}

\paragraph{Domain-validity guidance.}
Removing domain-validity reward gates from
SLIC-\textsc{Random} reduces Turn~8 Pass@1 by 4.3 percentage
points in-distribution and 7.0 points out-of-distribution
(Tables~\ref{tab:val_turns} and~\ref{tab:test_topo_turns}), which supports our inductive bias that bringing a circuit into a feasible operating region before further optimization is beneficial.

\newlength{\savedintextsep}
\setlength{\savedintextsep}{\intextsep}
\setlength{\intextsep}{0pt}
\begin{wraptable}{r}{0.43\textwidth}
\vspace{-0.6\baselineskip}
\centering
\small
\setlength{\tabcolsep}{4pt}
\renewcommand{\arraystretch}{1.08}
\caption{
Slack bonus ablation with random continuation and
dynamic filtering.
}
\label{tab:slack_bonus_ablation}
\begin{adjustbox}{max width=\linewidth}
\begin{tabular}{@{}lcc@{}}
\toprule
Slack bonus & ID & OOD \\
\midrule
Without
& $76.4 \pm 0.9$ & $56.7 \pm 1.3$ \\
With
& $84.9 \pm 0.7$ & $68.8 \pm 2.2$ \\
\bottomrule
\end{tabular}
\end{adjustbox}
\end{wraptable}

\paragraph{Success-conditional slack bonus.}
Table~\ref{tab:slack_bonus_ablation} reports four-specification
Pass@1 at Turn~8. With dynamic filtering enabled, adding the slack bonus
improves Pass@1 by 8.5 percentage points on ID tasks
and 12.1 points on held-out topologies.
The larger gain on held-out topologies supports the
benefit of this richer feedback for generalization.

Additional ablations in Appendix~\ref{app:additional_ablations}
examine advantage normalization, rollout construction,
episode termination, training horizon, and history
representation and length.
\par
\setlength{\intextsep}{\savedintextsep}

\subsection{{Backbone Generalization and Test-Time Scaling}}
\label{sec:backbone_generalization}

\paragraph{Generalization across backbones.}Table~\ref{tab:backbone_generalization} shows that
SLIC improves over SFT across all four
evaluated backbones.
Within the Qwen3.5 family, scaling from 4B to 9B improves
Pass@1 by 1.8 percentage points on ID tasks and
5.5 points on held-out topologies, suggesting additional benefits from
scaling model capacity.

\begin{table}[t]
\centering
\small
\setlength{\tabcolsep}{4pt}
\caption{
Generalization across backbones at Turn~8 on 840 ID and
240 OOD tasks.
Qwen3 and Qwen3.5 backbones follow the main
SLIC-\textsc{Random} configuration.
$^{\dagger}$Qwen2.5-7B-Instruct uses
SLIC-\textsc{Softmax} with 800 RL updates
(Appendix~\ref{app:qwen25_65nm}).
}
\label{tab:backbone_generalization}
\begin{adjustbox}{max width=\linewidth}
\begin{tabular}{@{}lcccccc@{}}
\toprule
& \multicolumn{3}{c}{In-distribution Pass@1 (\%)}
& \multicolumn{3}{c}{Out-of-distribution Pass@1 (\%)} \\
\cmidrule(lr){2-4}
\cmidrule(lr){5-7}
Backbone
& Base & SFT & SLIC
& Base & SFT & SLIC \\
\midrule
Qwen2.5-7B-Instruct$^{\dagger}$
& $1.0 \pm 0.2$
& $14.0 \pm 0.7$
& $32.2 \pm 0.8$
& $2.8 \pm 0.6$
& $13.3 \pm 1.1$
& $24.9 \pm 1.7$ \\
Qwen3-8B
& $1.4 \pm 0.0$
& $13.0 \pm 0.9$
& $39.8 \pm 0.5$
& $3.2 \pm 0.2$
& $11.4 \pm 1.3$
& $29.0 \pm 1.2$ \\
Qwen3.5-4B
& $2.3 \pm 0.5$
& $26.5 \pm 0.7$
& $87.1 \pm 0.7$
& $4.4 \pm 0.9$
& $19.9 \pm 1.3$
& $70.6 \pm 1.2$ \\
Qwen3.5-9B
& $3.0 \pm 0.2$
& $27.1 \pm 2.0$
& $88.9 \pm 0.4$
& $4.2 \pm 0.7$
& $16.9 \pm 1.0$
& $76.1 \pm 1.3$ \\
\bottomrule
\end{tabular}
\end{adjustbox}
\vspace{-1.2em}
\end{table}

\paragraph{Pass@1 over interaction turns.}
Figure~\ref{fig:val_pass_by_turn} shows that additional
test-time simulator interactions improve cumulative
Pass@1 on both splits, with the largest gains in early
turns and diminishing returns at longer horizons. This suggests that much of the benefit can be captured
within a moderate interaction budget, motivating us
to examine shorter training horizons.
Training with four turns remains effective and allows
continued improvement beyond the training horizon.
Further shortening the training horizon to single-turn GRPO lowers Pass@1
(Tables~\ref{tab:val_turns} and~\ref{tab:test_topo_turns})
and leads to earlier saturation, particularly on held-out topologies.
This is consistent with its lack of training experience
in successive simulator-guided refinement.
These results suggest that a moderate multi-turn training
budget may reduce training compute while preserving
test-time scaling (Appendix~\ref{app:additional_ablations}).

\begingroup
\setlength{\intextsep}{8pt}
\begin{figure}[h]
  \centering
\includegraphics[width=\linewidth]{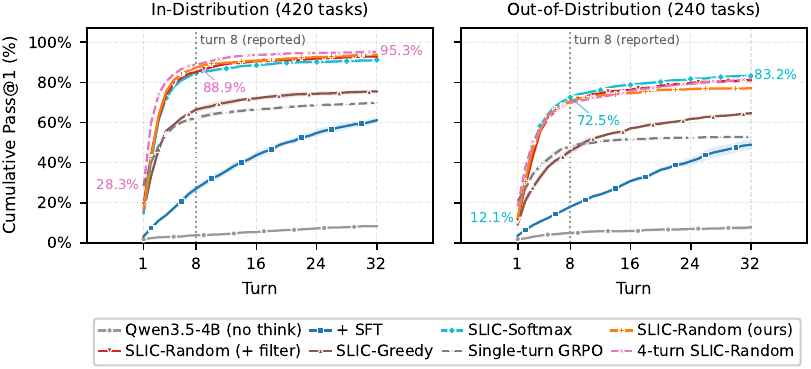}
  \vspace{-1.2em}
  \caption{
Test-time scaling of cumulative Pass@1 with up to 32
interaction turns on a 420-task subset of the full 840 ID tasks (left) and 240 OOD
tasks (right). Qwen3.5-4B is evaluated in non-thinking mode; we also tested
thinking mode with an 8k-token generation budget, but reasoning
did not terminate within this limit. Numerical annotations highlight selected values along
the best-performing curve in each panel.
}
  \label{fig:val_pass_by_turn}
\end{figure}
\endgroup

\begin{figure*}[t]
    \centering
    \includegraphics[width=\textwidth]{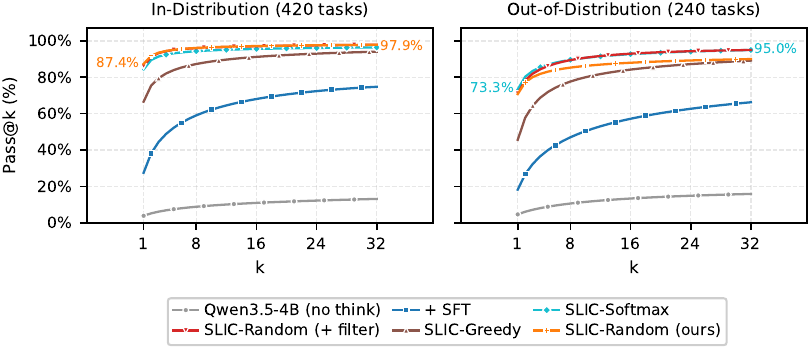}
    \vspace{-1.4em}
    \caption{
    Test-time scaling of Pass@$k$ with the number of sampled
rollouts on a 420-task subset of the full 840 ID tasks (left) and 240 OOD tasks (right) with
 eight-turn simulator budget per rollout. Numerical annotations highlight selected values along
the best-performing curve in each panel.
    }
    \label{fig:pass_at_k}
    \vspace{-1.2em}
\end{figure*}

\paragraph{Pass@$k$ over $k$.}
Increasing the number of test-time rollouts raises
Pass@$k$ toward saturation across the evaluated methods,
although SLIC's higher first-attempt success leaves
less room for improvement
(Figure~\ref{fig:pass_at_k}).
SFT attains substantially higher success rates
with a large sampling budget, suggesting that successful
optimization strategies are already accessible to the policy.
SLIC-\textsc{Random} exceeds SFT's Pass@32 with a single
rollout on both splits, consistent with reinforcement
fine-tuning concentrating probability mass on effective
optimization behaviors~\citep{yue2025doesrl} and reducing the need for
repeated sampling.

\section{Conclusion and Future Work}
We present SLIC, a simulator-interactive RLVR framework that formulates analog circuit optimization as a multi-turn language-model decision problem grounded in verifiable simulator feedback. With structured circuit representations, interpretable relative-update actions, domain-aware reward shaping, and reward-guided rollout steering, a 4B policy reaches 87.1\% in-distribution Pass@1 within eight turns, 3.5 percentage points below GPT-6-Astra, and exceeds GPT-6-Astra at four turns. The trained policies transfer to held-out topologies, five unseen PTM nodes, and a commercial TSMC 65\,nm PDK without additional fine-tuning, while running entirely with local inference. These results position simulator-grounded RLVR on small models as a practical path toward locally deployable analog circuit design automation.

Future work will extend this formulation toward the multi-agent analog design framework envisioned in the introduction. 
Rather than relying on a single design-optimization agent, future systems may coordinate multiple specialized agents for topology exploration, block-level optimization, verification, layout-aware refinement, and system-level specification decomposition. 
This direction better matches the modular and hierarchical nature of analog circuit design, but also raises new challenges in scaling to larger analog and mixed-signal systems, including long-context management, cross-agent coordination, and maintaining consistent design state across hierarchical blocks. 
Addressing these issues could enable scalable collaboration among lightweight, locally deployable agents for more complex circuit and system-level design workflows.

Additional experiments, implementation details, limitations, and broader impact discussions are provided in the Appendix.

\section*{Acknowledgments}
This work was supported primarily by Analog Devices (ADI),
with additional funding from the MIT Cadence Fellowship
and the JLO/MACOM Endowment Fund.
We thank the MIT Center for Integrated Circuits and
Systems (CICS) for its support.
We also acknowledge IBM, the MIT Office of Research
Computing and Data (ORCD), and the Artificial Intelligence
Compute Resources (AICR) cluster for providing computing
and storage resources.
{
\small
\bibliographystyle{plainnat-shortauthors}
\bibliography{references}
}
\clearpage

\appendix
\pdfbookmark[0]{Appendix}{slic-appendix}
\section{Related Work}
\paragraph{Analog circuit design and automation.}
Prior work on analog circuit design automation mainly follows two directions. The first treats sizing as black-box optimization~\citep{shahriari2015taking} over simulator feedback. Representative early efforts investigated single-objective~\citep{liu2013gaussian} or multi-objective Bayesian optimization~\citep{lyu2018batch,zhengqi2019bo} for analog circuit design. We refer readers to the review paper~\citep{gao2023review} for a comprehensive overview. Subsequent developments, such as TuRBO~\citep{eriksson2019turbo}, further improved the scalability of Bayesian optimization by using local trust regions. More recently, KATO \citep{kato2024} studies knowledge transfer for transistor sizing across different circuits and technology nodes. Despite their effectiveness, these approaches still optimize through implicit black-box search, offering limited interpretability and no explicit multi-turn reasoning process. The second direction learns optimization policies directly from data or simulation~\citep{jiang2026fdmagrpo, ahmadzadeh2025anacraft}. AutoCkt \citep{settaluri2020autockt} formulates analog design as a reinforcement learning problem over circuit environments. GCN-RL \citep{wang2020gcnrl} further introduces graph convolutional neural networks to improve transfer across technology nodes and similar circuit topologies. While these learning-based approaches demonstrate the promise of neural automation, they typically train a topology-specific optimizer or policy network rather than a general-purpose reasoning model.

\paragraph{Multi-turn and multi-reward reinforcement learning.}
Our method is closely related to recent work on reinforcement learning for multi-turn agentic optimization~\citep{wang2025practitioner,zhong2026rcgrpo}. MT-GRPO \citep{zeng2025mtgrpo} studies turn-level credit assignment in multi-turn reasoning trajectories, highlighting that purely trajectory-level returns can produce ambiguous supervision across turns~\citep{zhou2025sweetrl,xie2026tips}. GiGPO \citep{feng2025gigpo} further provides a practical formulation for multi-turn policy optimization over structured agent trajectories, and our implementation builds on this line of formulation. GTPO~\citep{ding2025gtpo} combines turn-level rewards
and return-based advantages for tool-integrated reasoning.
Complementary theoretical work analyzes the process-level
credit assignment implicit in outcome-based
GRPO~\citep{grpo_prm2025}. Our setting is also related to multi-reward reinforcement learning, where multiple objectives must be combined into a stable scalar training signal. GDPO \citep{liu2026gdpo} addresses this issue by introducing decoupled normalization for GRPO under multiple reward channels. In addition, Hindsight Experience Replay (HER) \citep{andrychowicz2017her} shows that achieved outcomes can be relabeled as training targets in goal-conditioned reinforcement learning, which is closely related to our automatic goal construction strategy. 
Concurrent work TL-GRPO~\citep{li2026tlgrpo} also studies turn-level GRPO-style reinforcement learning for analog sizing. In contrast, we focus on problem-driven, locally deployable small-model training, with domain adaptations for circuit representation, multi-objective rewards, reward-guided rollout sampling, and simulator-interactive evaluation across topologies and technology nodes.

\paragraph{Large language models for circuit design.}
Recent work has begun to explore LLMs for analog design automation~\citep{yin2024adollm,shen2025atelier,wu2025easysize,somayaji2025llmuso,shen2026vlmcad}. AnalogCoder~\citep{lai2024analogcoder} studies training-free analog circuit design through Python code generation with feedback-enhanced prompting. EEsizer~\citep{liu2025eesizer} integrates an LLM with circuit simulation and custom analysis functions for closed-loop transistor sizing. LEDRO~\citep{kochar2025ledro} uses LLMs to iteratively reduce the analog sizing search space and guide downstream optimization, with an emphasis on cross-topology and cross-technology generalization. AutoSizer~\citep{yu2026autosizer} further proposes a reflective two-loop meta-optimization framework and introduces AMS-SizingBench for evaluation. These works introduce stronger knowledge priors and often improve interpretability over earlier black-box methods, but they primarily rely on prompting, reflection, or agent scaffolds around fixed foundation models. While prior work uses reinforcement learning to train an LLM
for circuit topology generation~\citep{vijayaraghavan2025autocircuitrl}, we focus on training lightweight, locally deployable SLM agents for closed-loop analog circuit optimization, which is important for realistic workflows where proprietary PDKs and simulator models cannot be exposed to external APIs. This shifts the emphasis from using large hosted models as design assistants to adapting smaller models themselves for simulator-grounded multi-turn optimization.

\section{Circuit Representation and Action Format}
\label{app:representation}

\subsection{XML Circuit Representation}
We represent circuits using an XML-like structured text format 
augmented with explicit handle annotations. Unlike raw SPICE 
netlists, which are designed for simulator compatibility and often 
encode many task-relevant relations only implicitly, this 
representation makes component types, editable attributes, and 
structural boundaries explicit in the token stream. This is 
particularly useful in our multi-turn setting, where the model 
must repeatedly inspect the current circuit state and produce 
targeted edits. The handle annotations further provide stable 
references for editable circuit elements across turns, reducing 
ambiguity during iterative optimization. A representative prompt excerpt is shown in the case study (Appendix~\ref{app:case}).

\subsection{Discretized Relative Actions}
The policy outputs each circuit edit action as a tuple (\textit{target component}, 
\textit{parameter}, \textit{direction}, \textit{ratio}), which the 
simulator applies as a multiplicative update to the current 
parameter value rather than generating an absolute new value from 
scratch. The framework does not prescribe a fixed number of circuit parameter edits per turn; instead, the model may generate a variable-length sequence of circuit edit actions in each single turn. For example, \texttt{increase @27 l\_MM4 50\%} multiplies the current channel length of the device at handle \texttt{@27} by 
$1.5$. The ratio is a positive integer percentage with a configurable upper bound; we cap it at $500\%$ in our experiments, which bounds the per-action perturbation while 
still allowing coarse early-stage circuit edits. This converts continuous parameter tuning into a finite set 
of local refinement operations and matches the iterative nature of 
analog sizing: each action refines the current design state instead 
of respecifying the full parameter configuration. The relative 
formulation is also less sensitive to scale variation across 
devices, topologies, and technology nodes.
\subsection{Action Validity}
\label{app:action_validity}

SLIC enforces a multi-layer validity check on every action, with 
three pre-invocation filters and one post-execution detector. 
(1) \textit{Format check}: actions failing to parse against the 
relative-update template are rejected. (2) \textit{Admissibility 
check}: actions referencing components or parameters outside the 
dynamically generated admissible list are rejected. (3) 
\textit{Parameter Range check}: after applying the multiplicative update, 
parameters falling outside their node-specific valid ranges (Table~\ref{tab:tech_node_ranges}) are 
rejected. (4) \textit{Simulator failure}: actions that pass the 
three pre-invocation filters but trigger DC/AC convergence 
failures or non-finite measurements during simulation are treated 
as invalid post-execution. All four cases incur the 
invalid-transition penalty $\lambda_{\mathrm{inv}}$ and contribute 
to GRPO advantages under joint normalization. Under separate normalization, invalidity affects the
validity advantage, while candidates without usable
performance measurements are excluded from the
performance comparison. The first three additionally save simulator 
budget by rejecting actions before any simulator call.

\begin{table}[H]
\centering
\caption{
Technology-specific valid ranges for editable transistor parameters
used in the parameter range check
(Appendix~\ref{app:action_validity}, item~(3)).
}
\label{tab:tech_node_ranges}
\small
\begin{tabular}{@{}lcc@{}}
\toprule
Technology & Channel length $L$ & Channel width $W$ \\
\midrule
PTM 22\,nm
& 22\,nm--100\,$\mu$m & 45\,nm--500\,$\mu$m \\
PTM 32\,nm
& 45\,nm--100\,$\mu$m & 45\,nm--500\,$\mu$m \\
PTM 45\,nm
& 45\,nm--100\,$\mu$m & 45\,nm--500\,$\mu$m \\
PTM 65\,nm (training)$^{\dagger}$
& 45\,nm--100\,$\mu$m & 45\,nm--500\,$\mu$m \\
PTM 90\,nm
& 56\,nm--100\,$\mu$m & 45\,nm--500\,$\mu$m \\
PTM 130\,nm
& 85\,nm--100\,$\mu$m & 45\,nm--500\,$\mu$m \\
TSMC CMOS 65\,nm
& 60\,nm--20\,$\mu$m & 120\,nm--500\,$\mu$m \\
\bottomrule
\end{tabular}

\smallskip
\begin{minipage}{0.95\linewidth}
\footnotesize
$^{\dagger}$At PTM 65\,nm, nine training topologies have
topology-specific upper bounds on $L$ ranging from
2.5 to 40\,$\mu$m. No held-out topology is affected.
\end{minipage}
\end{table}

\section{Reward Function Details}
\label{app:reward_details}
\subsection{Per-Specification Reward}
We consider four specifications indexed by
$\mathcal{K}=\{\Gamma,\mathrm{B},\mathrm{P},\varphi\}$,
corresponding to gain, bandwidth, power consumption,
and phase margin.
At turn $t$, their measured values are
\[
m_t^{(\Gamma)}=\Gamma_t,\qquad
m_t^{(\mathrm{B})}=B_t,\qquad
m_t^{(\mathrm{P})}=P_t,\qquad
m_t^{(\varphi)}=\varphi_t.
\]
These measurements are obtained after executing the
candidate action.
We omit the candidate index when describing a single
reward evaluation.
Gain is expressed in dB.

For gain, bandwidth, and power, let $q^{(k)}$ denote
the corresponding target in the task specification $q$,
with $\Gamma^\star=q^{(\Gamma)}$.
We define
\begin{equation}
r_t^{(k)}
=
\min\!\left(
1,
\left(
\frac{m_t^{(k)}+\delta}{q^{(k)}+\delta}
\right)^{\eta_k}
\right),
\qquad
\eta_k=
\begin{cases}
+1, & k\in\{\Gamma,\mathrm{B}\},\\
-1, & k=\mathrm{P}.
\end{cases}
\label{eq:unified_spec_reward}
\end{equation}
The direction $\eta_k$ reflects that higher gain and
bandwidth are preferred, whereas lower power is preferred.
When $\delta$ is negligible, the scores approach
$m_t^{(k)}/q^{(k)}$ for gain and bandwidth and
$q^{(k)}/m_t^{(k)}$ for power, capped at one.
The small constant $\delta>0$ prevents division by zero
for nonnegative measurements and targets.

For a finite phase-margin measurement, the reward is
the indicator of whether it lies within the target
window $[\varphi_{\mathrm{lo}},\varphi_{\mathrm{hi}}]$:
\begin{equation}
r_t^{(\varphi)}
=
b_t^{(\varphi)}
=
\mathbb{I}\!\left[
\varphi_{\mathrm{lo}}\leq\varphi_t\leq\varphi_{\mathrm{hi}}
\right],
\label{eq:bounded_spec_check}
\end{equation}
where $\mathbb{I}[\cdot]$ is the indicator function.
The no-unity-gain-crossing case is handled separately
under the domain-validity rules.

\subsection{Domain-Validity Gates}

Let $F_t$ denote the in-band peaking measure in dB.
We define $d_t=\mathbb{I}[\Gamma_t>0]$ to indicate positive gain and
$\chi_t=\mathbb{I}[F_t\text{ is finite and }F_t>6\,\mathrm{dB}]$
to indicate excessive peaking.
The gated per-specification rewards are

\begin{equation}
\widetilde r_t^{(k)}=r_t^{(k)}
\begin{cases}
1-\chi_t, & k=\Gamma,\\
d_t, & k=\mathrm P,\\
d_t(1-\chi_t), & k\in\{\mathrm B,\varphi\}.
\end{cases}
\label{eq:domain_validity_gate}
\end{equation}
Non-positive gain therefore suppresses auxiliary performance
credit, while excessive peaking suppresses gain, bandwidth,
and phase-margin credit.
Power remains eligible when gain is positive because it is
measured from the DC operating point.
For a simulator-reported no-unity-gain-crossing case,
phase-margin credit requires positive gain and a finite
flatness measurement satisfying $F_t\leq6\,\mathrm{dB}$.

The same gates apply to specification satisfaction.
A design succeeds only when all target specifications pass
their tolerance-relaxed thresholds under these conditions.
We use relative tolerance $\varepsilon_{\mathrm{tol},k}=0.05$:
gain and bandwidth lower bounds are multiplied by $1-\varepsilon_{\mathrm{tol},k}$,
the power upper bound by $1+\varepsilon_{\mathrm{tol},k}$, and the phase-margin
window is widened about its center by a factor of $1+\varepsilon_{\mathrm{tol},\varphi}$.
Flatness is a validity condition rather than an additional
optimization objective.

\subsection{Success-Conditional Slack Bonus}

Uniform rewards for successful candidates can eliminate
group-relative learning signals, motivating the dynamic
sampling and filtering strategy in DAPO~\citep{yu2025dapo}.
In circuit optimization, successful candidates can still differ
in specification headroom.
We therefore retain this information through a bounded
success-conditional slack bonus.

For gain, bandwidth, and power, we define relative slack as
\begin{equation}
u_t^{(k)}
=
\eta_k
\frac{m_t^{(k)}-q^{(k)}}{|q^{(k)}|},
\qquad
k\in\{\Gamma,\mathrm{B},\mathrm{P}\},
\label{eq:relative_slack}
\end{equation}
using the optimization direction $\eta_k$ defined in
Equation~\ref{eq:unified_spec_reward}.
Positive slack indicates performance beyond the target
in the desired direction.

We map each slack to $[0,1]$ before averaging, assigning zero
at the tolerance-relaxed boundary and one at the prescribed
headroom cap.
For phase margin, the score instead measures proximity to
the center of the accepted window.
Writing $\varphi_c=(\varphi_{\mathrm{lo}}+\varphi_{\mathrm{hi}})/2$
and $h_\varphi=(\varphi_{\mathrm{hi}}-\varphi_{\mathrm{lo}})/2$, we define
\begin{equation}
\begin{aligned}
z_t^{(k)}
&=
\operatorname{clip}
\left(
\frac{u_t^{(k)}+\varepsilon_{\mathrm{tol},k}}{u_{\max}+\varepsilon_{\mathrm{tol},k}},0,1
\right),
&& k\in\{\Gamma,\mathrm{B},\mathrm{P}\},\quad u_{\max}=1,\\
z_t^{(\varphi)}
&=
\operatorname{clip}
\left(
1-\frac{|\varphi_t-\varphi_c|}
         {h_\varphi(1+\varepsilon_{\mathrm{tol},\varphi})},0,1
\right).
\end{aligned}
\label{eq:normalized_slack}
\end{equation}
The tolerance shift preserves distinctions between designs
that succeed within the tolerance band, while $u_{\max}=1$ saturates
the one-sided terms at $100\%$ relative headroom.
The phase-margin term rewards centering within the accepted window.
For a no-crossing case satisfying the domain-validity conditions,
we set $z_t^{(\varphi)}=1$.

\subsection{Separate Normalization of Validity and Performance}
\label{app:separate_normalization}
Following the reward-decoupled normalization principle of
GDPO~\citep{liu2026gdpo}, we normalize action validity and measured
performance separately before combining their advantages.
In mixed-validity groups, the separation between valid actions
and constant-penalty failures can dominate aggregate reward
variance, compressing performance contrast among measured designs.

At turn $t$, let $\mathcal{I}_t=\{1,\ldots,G\}$ index the
$G$ candidate actions, and let $v_{t,i}\in\{0,1\}$ denote
the action-validity label of candidate $i$.
Let $\mathcal{I}_t^{\mathrm{meas}}\subseteq\mathcal{I}_t$ contain candidates
for which the simulator returns sufficient measurements
to compute a performance reward $r_{t,i}$.

We compute a validity advantage relative to the full candidate
group and a performance advantage relative to the candidates
with available performance rewards:
\begin{equation}
\begin{aligned}
A_{t,i}^{\mathrm{val}}
&= \mathcal{N}_{\mathcal{I}_t}(v_t)_i,\\
A_{t,i}^{\mathrm{perf}}
&=
\begin{cases}
\mathcal{N}_{\mathcal{I}_t^{\mathrm{meas}}}(\mathbf r_t)_i,
    & i\in\mathcal{I}_t^{\mathrm{meas}},\\
0, & i\notin\mathcal{I}_t^{\mathrm{meas}}.
\end{cases}
\end{aligned}
\label{eq:separate_normalization}
\end{equation}
Here, $v_t$ and $\mathbf r_t$ collect the validity labels and
performance rewards at turn $t$, respectively.
For a comparison set $\mathcal{S}$ with at least two
members and nonzero variance, the normalization operator is
\[
\mathcal{N}_{\mathcal{S}}(x)_i
=
\frac{x_i-\mu_{\mathcal{S}}(x)}
     {\operatorname{std}_{\mathcal{S}}(x)+\epsilon},
\]
where $\mu_{\mathcal{S}}(x)$ and
$\operatorname{std}_{\mathcal{S}}(x)$ are the mean and
standard deviation of $\{x_j:j\in\mathcal{S}\}$,
and $\epsilon>0$ ensures numerical stability. Empty or singleton comparison sets and sets with zero
variance contribute zero advantage.
Thus, each advantage measures how a candidate compares
with the mean of its respective comparison set.

The combined turn-level advantage is
\begin{equation}
A_{t,i}
=
\lambda_{\mathrm{val}}A_{t,i}^{\mathrm{val}}
+
\lambda_{\mathrm{perf}}A_{t,i}^{\mathrm{perf}},
\label{eq:combined_advantage}
\end{equation}
where $\lambda_{\mathrm{val}}$ and $\lambda_{\mathrm{perf}}$
weight the validity and performance advantages, respectively.
We set $\lambda_{\mathrm{val}}=\lambda_{\mathrm{perf}}=1$. This separates validity discrimination from performance ranking
without interpreting unavailable measurements as ordinary
low-performance observations.
The combined advantage is assigned to the generated response tokens, including reasoning and actions, and used in the clipped policy objective.

\subsection{Final Reward}

The final turn reward is
\begin{equation}
r_t=
\begin{cases}
\lambda_{\mathrm{inv}},
& \text{no usable performance measurement},\\[1mm]
\displaystyle
1+\frac{w}{|\mathcal{K}|}
\sum_{k\in\mathcal{K}}z_t^{(k)},
& \text{design succeeds within tolerance},\\[3mm]
10^{(\Gamma_t-\Gamma^\star)/20},
& \Gamma_t<0,\\[1mm]
\displaystyle
\min\left(
1-\epsilon_s,\,
\frac{1}{|\mathcal{K}|}
\sum_{k\in\mathcal{K}}\widetilde r_t^{(k)}
\right),
& \text{otherwise},
\end{cases}
\label{eq:final_reward}
\end{equation}
where $w=0.5$, $\lambda_{\mathrm{inv}}=-0.1$, and
$\epsilon_s>0$ caps unsuccessful scores strictly below one.
Successful scores lie in $[1,1.5]$, with the bonus reflecting
mean specification headroom.
Under separate normalization, $\lambda_{\mathrm{inv}}$
does not enter the advantage calculation.
The same reward applies whether episodes stop at success
or continue optimizing afterward.

\section{Reward-Guided Rollout Steering Algorithm}
\label{app:rgrs_algo}

At each turn $t$, the policy samples $G$ candidate edits 
$a_{t,i} \sim \pi_\theta(\cdot \mid h_t)$ from the shared history 
$h_t$. Here, $h_t$ denotes the prompt containing the task specification, current circuit state, and retained interaction history. The simulator executes each candidate, producing successor 
states $s_{t+1,i}$, validity labels $v_{t,i} \in \{0,1\}$, and 
post-action rewards $r_{t,i}$ from Section~\ref{sec:reward}. We 
then form the eligible set
\begin{equation}
\mathcal{E}_t =
\begin{cases}
\{i : v_{t,i}=1\}, & \text{if nonempty},\\
\{1,\dots,G\}, & \text{otherwise (random fallback)},
\end{cases}
\label{eq:eligible_candidate_set}
\end{equation}
and a continuation policy $\kappa$ selects 
$i_t^\star \sim \kappa(\{r_{t,i}\}_{i \in \mathcal{E}_t})$ to 
advance the shared history. All $G$ transitions are retained for 
group-relative policy updates. For softmax continuation, $\tau_{\mathrm{sel}}>0$
    controls the preference for higher-scoring candidates. The full algorithm is provided in Algorithm~\ref{alg:reward_guided_continuation}.

\vspace{4pt}
\begin{algorithm}[H]
\caption{Reward-Guided Rollout Steering (RGRS)}
\label{alg:reward_guided_continuation}
\begin{algorithmic}[1]
\REQUIRE Initial circuit state $s_0$, target specification $q$, policy $\pi_\theta$, horizon $T$, group size $G$, continuation policy $\kappa$
\STATE Initialize shared history $h_0 \leftarrow \mathrm{Prompt}(q, s_0)$
\STATE Initialize buffer $\mathcal{D} \leftarrow \emptyset$
\FOR{$t=0,\ldots,T-1$}
    \STATE Sample candidate edits $\{a_{t,i}\}_{i=1}^{G}$ from $\pi_\theta(\cdot \mid h_t)$
    \FOR{$i=1,\ldots,G$}
        \STATE Execute $a_{t,i}$ in the simulator to obtain successor state $s_{t+1,i}$
        \STATE Compute validity label $v_{t,i}$ and post-action reward $r_{t,i}$
        \STATE Store $(h_t, a_{t,i}, s_{t+1,i}, r_{t,i}, v_{t,i})$ in $\mathcal{D}$
    \ENDFOR
    \STATE Construct valid set $\mathcal{V}_t \leftarrow \{i \in \{1,\ldots,G\}: v_{t,i}=1\}$
    \IF{$\mathcal{V}_t=\emptyset$}
        \STATE Set eligible set $\mathcal{E}_t \leftarrow \{1,\ldots,G\}$
    \ELSE
        \STATE Set eligible set $\mathcal{E}_t \leftarrow \mathcal{V}_t$
    \ENDIF
    \IF{$\kappa=\textsc{Random}$ or $\mathcal{V}_t=\emptyset$}
        \STATE Sample $i_t^\star$ uniformly from $\mathcal{E}_t$
    \ELSIF{$\kappa=\textsc{Greedy}$}
        \STATE Set $i_t^\star \leftarrow \arg\max_{i\in\mathcal{E}_t} r_{t,i}$
    \ELSIF{$\kappa=\textsc{Softmax}$}
        \STATE Sample $i_t^\star$ from $\mathcal{E}_t$ with $p(i)\propto \exp(r_{t,i}/\tau_{\mathrm{sel}})$
    \ENDIF
    \STATE Update shared history $h_{t+1} \leftarrow \mathrm{Update}(h_t, a_{t,i_t^\star}, s_{t+1,i_t^\star})$
    \IF{$s_{t+1,i_t^\star}$ satisfies all target specifications}
        \STATE \textbf{break}
    \ENDIF
\ENDFOR
 \STATE Compute turn-level advantages as in Appendix~\ref{app:separate_normalization} and update
    $\pi_\theta$ using all transitions in $\mathcal D$
\end{algorithmic}
\end{algorithm}

\section{Additional Ablations}
\label{app:additional_ablations}

We report four-specification Pass@1 (\%) at Turn~8 on
840 in-distribution and 240 out-of-distribution tasks.
Each comparison changes only the indicated component within
its specified training setting.
All variants use the same evaluation targets and
domain-validity criteria.

\subsection{Rollout Construction}

\begin{table}[t]
\centering

\begin{minipage}[t]{0.48\linewidth}
\vspace{0pt}
\caption{
Trajectory construction with dynamic filtering,
no slack bonus, and stopping at success.
RGRS uses random continuation.
Turn~8 Pass@1 (\%).
}
\label{tab:trajectory_construction_ablation}
\end{minipage}
\hfill
\begin{minipage}[t]{0.48\linewidth}
\vspace{0pt}
\caption{
Continuation candidate set with softmax selection,
joint normalization, and continued optimization after success.
Turn~8 Pass@1 (\%).
}
\label{tab:candidate_set_ablation}
\end{minipage}

\par\vspace{2pt}

\begin{minipage}[t]{0.48\linewidth}
\vspace{0pt}
\centering
\small
\setlength{\tabcolsep}{3pt}
\renewcommand{\arraystretch}{1.08}
\begin{tabular*}{\linewidth}{@{}l@{\extracolsep{\fill}}cc@{}}
\toprule
Construction
& \shortstack{In-\\distribution}
& \shortstack{Out-of-\\distribution} \\
\midrule
Fixed-base (TL-GRPO)
& $73.8 \pm 0.5$
& $57.6 \pm 0.6$ \\
Online (RGRS)
& $76.4 \pm 0.9$
& $56.7 \pm 1.3$ \\
\bottomrule
\end{tabular*}
\end{minipage}
\hfill
\begin{minipage}[t]{0.48\linewidth}
\vspace{0pt}
\centering
\small
\setlength{\tabcolsep}{3pt}
\renewcommand{\arraystretch}{1.08}
\begin{tabular*}{\linewidth}{@{}l@{\extracolsep{\fill}}cc@{}}
\toprule
Candidate set
& \shortstack{In-\\distribution}
& \shortstack{Out-of-\\distribution} \\
\midrule
Eligible only
& $77.5 \pm 0.3$
& $59.3 \pm 1.3$ \\
All candidates
& $76.9 \pm 0.4$
& $59.6 \pm 0.7$ \\
\bottomrule
\end{tabular*}
\end{minipage}
\end{table}

\paragraph{Trajectory construction.}
With dynamic filtering~\citep{yu2025dapo}
enabled and the slack bonus disabled,
online construction improves in-distribution Pass@1 by
2.6 percentage points over fixed-base TL-GRPO~\citep{li2026tlgrpo}, while
out-of-distribution Pass@1 decreases by 0.9 points
(Table~\ref{tab:trajectory_construction_ablation}).
Thus, this comparison shows an in-distribution benefit but
does not establish a generalization advantage for
online construction alone.

\paragraph{Continuation candidate set.}
Restricting softmax selection to eligible candidates changes
in-distribution and out-of-distribution Pass@1 by
$+0.6$ and $-0.3$ percentage points, respectively
(Table~\ref{tab:candidate_set_ablation}).
The small, mixed differences suggest limited sensitivity
to this restriction.

\subsection{Advantage Normalization}
\label{app:normalization_ablation}

With episodes continuing after success, separate normalization
improves in-distribution and out-of-distribution Pass@1 by
6.2 and 4.9 percentage points, respectively
(Table~\ref{tab:normalization_ablation}).
Under joint normalization, the reward gap between invalid actions
and measured designs can inflate group variance, compressing
advantage differences among designs with distinct performance.
Normalizing validity and performance separately preserves these
comparisons and avoids using the invalid-action penalty to
calibrate the scale of performance advantages.

\begin{table}[t]
\centering
\small
\setlength{\tabcolsep}{4pt}
\renewcommand{\arraystretch}{1.08}

\begin{minipage}[t]{0.48\linewidth}
\vspace{0pt}
\centering
\caption{
Normalization ablation for Qwen3.5-4B with random continuation and
continued optimization after success.
Results are four-specification Pass@1
(\%) at Turn~8.
}
\label{tab:normalization_ablation}
\begin{adjustbox}{max width=\linewidth}
\begin{tabular}{@{}lcc@{}}
\toprule
Normalization & In-distribution & Out-of-distribution \\
\midrule
Joint
& $79.0 \pm 0.7$ & $60.4 \pm 2.3$ \\
Separate
& $85.2 \pm 2.2$ & $65.3 \pm 2.4$ \\
\bottomrule
\end{tabular}
\end{adjustbox}
\end{minipage}
\hfill
\begin{minipage}[t]{0.48\linewidth}
\vspace{0pt}
\centering
\caption{
History-length ablation for Qwen3.5-4B with 2k training tasks,
softmax continuation, and joint normalization.
Results are four-specification Pass@1
(\%) at Turn~8.
}
\label{tab:history_length_ablation}
\begin{adjustbox}{max width=\linewidth}
\begin{tabular}{@{}lcc@{}}
\toprule
History length & In-distribution & Out-of-distribution \\
\midrule
$H=1$
& $82.8 \pm 0.7$ & $64.9 \pm 2.1$ \\
$H=8$
& $83.6 \pm 0.8$ & $69.2 \pm 2.5$ \\
\bottomrule
\end{tabular}
\end{adjustbox}
\end{minipage}

\end{table}

\subsection{Training Configuration and History}

\begin{table}[t]
\centering
\small
\setlength{\tabcolsep}{5pt}
\caption{
Ablations relative to the main SLIC-\textsc{Random} configuration,
which uses separate normalization, the success-conditional
slack bonus, stopping at success, an eight-turn training budget,
and structured history.
All policies are evaluated at Turn~8.
}
\label{tab:training_configuration_ablations}
\begin{adjustbox}{max width=\linewidth}
\begin{tabular}{@{}lcc@{}}
\toprule
Training configuration
& \shortstack{In-distribution}
& \shortstack{Out-of-distribution} \\
\midrule
Main configuration
& $87.1 \pm 0.7$ & $70.6 \pm 1.2$ \\
Continue optimization after success
& $85.2 \pm 2.2$ & $65.3 \pm 2.4$ \\
Single-turn GRPO
& $61.6 \pm 1.4$ & $47.6 \pm 0.9$ \\
Four-turn SLIC-\textsc{Random}
& $89.5 \pm 0.1$ & $69.3 \pm 0.6$ \\
Retain raw model outputs in history
& $76.8 \pm 0.5$ & $55.7 \pm 1.7$ \\
Train with zero tolerance
& $81.7 \pm 0.5$ & $60.1 \pm 0.6$ \\
\bottomrule
\end{tabular}
\end{adjustbox}
\end{table}

\paragraph{Termination after success.}
Stopping at success yields higher mean Pass@1 on both splits,
with a larger improvement on held-out topologies
(Table~\ref{tab:training_configuration_ablations}).
Continuing after success is better suited to searching
for higher-performing designs beyond target satisfaction,
while stopping at success fits our goal-directed setting.

\paragraph{Training horizon.}
Training with four turns remains competitive with the
main eight-turn configuration, whereas single-turn
training substantially reduces performance on both splits
when evaluated at eight turns
(Table~\ref{tab:training_configuration_ablations}).
For the main policy, rollouts that succeed within eight
evaluation turns reach their first success after an
average of 2.87 turns in-distribution and 3.16 turns
out-of-distribution.
This concentration of successful refinement within
a few turns is consistent with the effectiveness
of four-turn training.

Four-turn training also preserves the ability to improve
through additional test-time simulator feedback.
In contrast, single-turn training exhibits more pronounced
diminishing returns from additional evaluation turns
and saturates earlier, particularly on held-out topologies.
A single-turn budget provides no RL experience of
refinement through successive updates, which may limit
the ability to sustain improvement during evaluation.
Together, these observations suggest a trade-off between
training and test-time compute.
A moderate training horizon may reduce training cost
while retaining the iterative refinement behavior needed
to benefit from further simulator interactions.
\paragraph{History representation.}
The main configuration retains the parsed action and
simulator observation from the most recent turn ($H=1$).
The raw-history variant also uses $H=1$ but retains the
full model response, including reasoning and actions,
along with the simulator observation.
This change reduces performance on both splits
(Table~\ref{tab:training_configuration_ablations}).
Both variants start from the same SFT checkpoint, without
additional SFT for the raw-history format.
The degradation may partly reflect this format mismatch,
as well as greater difficulty identifying relevant actions
and simulator feedback within the additional reasoning text.

\paragraph{History length.}
We study retained history length using SLIC-\textsc{Softmax}
trained on a smaller dataset of 2,000 tasks, with joint
normalization held fixed.
Increasing history length from $H=1$ to $H=8$ yields
a small in-distribution improvement and a larger gain
on held-out topologies
(Table~\ref{tab:history_length_ablation}).
A longer history may help the policy compare earlier
parameter edits and their outcomes, guiding subsequent
exploration on unfamiliar circuit structures.
Both variants start from the same SFT checkpoint trained
with $H=1$.
Mean training prompt lengths with $H=8$ are approximately
5.8--6.6k tokens, exceeding the 4.6--4.8k tokens observed
in the raw-history experiment.
These results suggest that longer inputs need not impair
small-model performance when they provide useful,
structured interaction history.

\paragraph{Training tolerance.}
Using zero tolerance during training reduces Turn~8
in-distribution and out-of-distribution Pass@1 by
5.4 and 10.5 percentage points, respectively,
under the same evaluation criterion
(Table~\ref{tab:training_configuration_ablations}).
This suggests that tolerance-relaxed training thresholds
provide a more effective learning signal near specification
boundaries, whereas strict thresholds may place excessive
emphasis on small residual violations.

\subsection{Tolerance Sensitivity}
\label{app:tolerance}

\begin{figure*}[h]
    \centering
    \includegraphics[width=\textwidth]{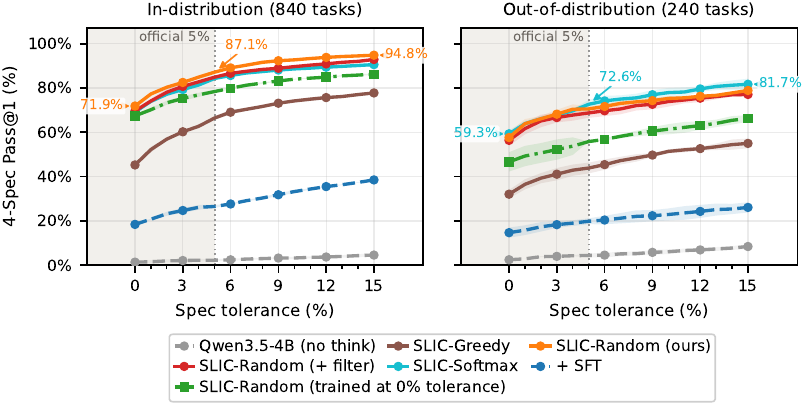}
\caption{
Tolerance sweep of four-specification Pass@1 within eight
turns on 840 in-distribution tasks (left) and 240
out-of-distribution tasks (right), with evaluation
tolerances ranging from 0\% to 15\%.
The vertical dotted line marks the default 5\% tolerance.
Trajectories are generated with strict targets and reevaluated
at each tolerance. Numerical annotations highlight selected values along
the best-performing curve in each panel.
}
    \label{fig:tolerance_sweep_val}
\end{figure*}

SLIC-\textsc{Random} trained with the default 5\%
success tolerance outperforms its zero-tolerance training
counterpart across the evaluated range, even when evaluation
requires strict target satisfaction (Figure~\ref{fig:tolerance_sweep_val}).
This suggests that modestly relaxing the training success
threshold can improve learning and generalization.
One possible explanation is that relaxing the success
threshold makes higher rewards more accessible early in
training, while the success-conditional slack bonus
continues to encourage specification headroom.
This may improve the utilization of sampled simulator
evaluations and help the policy progress toward strict
target satisfaction.

\section{Baseline Tree MT-GRPO Implementation}
\label{app:baseline_details}

\subsection{Tree MT-GRPO}
\label{app:tree_mt_grpo}

We implement Tree MT-GRPO~\citep{zeng2025mtgrpo} with
16 tasks per training step, matching the main
SLIC-\textsc{Random} run. To keep the simulation budgets
comparable, we use a four-turn training horizon and group
size $G=3$. Each active state branches into three candidate
actions in the first three turns, and each surviving branch
takes one action in the final turn. A fully expanded tree
thus contains $3+9+27+27=66$ candidate actions per task,
close to the $8\times8=64$ candidates in a full
SLIC-\textsc{Random} rollout. Episodes terminate at first success. All other optimization and generation settings follow the
main SLIC-\textsc{Random} configuration. Both methods are
evaluated with the standard eight-turn interaction budget.

Intermediate rewards use SLIC's performance score
for the updated circuit, with $-0.1$ assigned to
invalid actions.
The outcome reward evaluates the terminal circuit
and includes the success-conditional slack bonus
for successful designs.
We normalize intermediate rewards among sibling actions
to obtain $A^{\mathrm{int}}$, and outcome rewards across
trajectories for the same task to obtain $A^{\mathrm{out}}$.

Following MT-GRPO~\citep[Eq.~5]{zeng2025mtgrpo},
we combine current and future intermediate advantages
with the outcome advantage. With unit discount, the advantage at turn $t$ of
trajectory $i$, whose length is $T_i$, is
\begin{equation}
A_{t,i}^{\mathrm{tree}}
=
\sum_{t'=t}^{T_i-2} A_{t',i}^{\mathrm{int}}
+
A_i^{\mathrm{out}}.
\label{eq:tree_mt_grpo_advantage}
\end{equation}
The terminal turn $t=T_i-1$ receives only
the outcome advantage $A_i^{\mathrm{out}}$.
For a shared response node $n$ at turn $t$,
Eq.~\ref{eq:tree_mt_grpo_advantage} gives a trajectory
advantage $A_{t,i}^{\mathrm{tree}}$ for each continuation $i$.
We average these advantages to obtain a single training
advantage $\overline A(n)$.
With $\operatorname{Ch}(n)$ denoting its child responses,
\begin{equation}
\begin{aligned}
\overline A(n)
&=
\mathbb{E}_{i\mid n}
\left[A_{t,i}^{\mathrm{tree}}\right]\\
&=
\begin{cases}
A^{\mathrm{out}}(n),
& \text{if $n$ is terminal},\\[1mm]
A^{\mathrm{int}}(n)
+
\displaystyle
\operatorname{mean}_{n'\in\operatorname{Ch}(n)}
\overline A(n'),
& \text{otherwise}.
\end{cases}
\end{aligned}
\label{eq:tree_shared_prefix_advantage}
\end{equation}
The expectation assigns equal weight to each child
    branch recursively.
At each non-terminal node, we add its intermediate
advantage to the mean of its children's advantages,
which already include all subsequent intermediate
and outcome terms.
Each shared response is trained once, with
$\overline A(n)$ assigned to all its generated tokens.

Tree depth is the main scalability bottleneck. At $G=3$,
extending the fully expanded tree from four to eight turns
increases the number of trajectories per task from $3^3=27$
to $3^7=2{,}187$. At eight turns, increasing the group size
from three to four raises this count to $4^7=16{,}384$. Under a fixed simulation budget, this exponential growth
makes it difficult to scale tree-based training to more complex circuits and tasks
that require more successive refinement steps to reach a successful design.
\section{Training and Evaluation with Qwen2.5-7B-Instruct}
\label{app:qwen25_evaluation}

\subsection{Three-Specification Setting at 22~nm}
\label{app:original_evaluation}

We evaluate an earlier configuration that trains Qwen2.5-7B-Instruct at the 22\,nm PTM node with a four-turn horizon and optimizes gain, power consumption, and phase margin. Evaluation covers 196 in-distribution tasks across 31 training topologies and 246 tasks across three held-out topologies at four and eight turns. The trajectory-level GRPO baseline uses independently
sampled episodes with terminal-return supervision
and trains stably in this three-specification setting.
\begin{table*}[t]
\centering
\small
\setlength{\tabcolsep}{7pt}
\renewcommand{\arraystretch}{1.12}
\caption{
Three-specification setting with Qwen2.5-7B-Instruct at 22\,nm,
evaluated on 196 in-distribution and 246 held-out-topology tasks.
All trained policies use a four-turn training horizon.
Bold denotes the highest mean in each column.
}
\label{tab:original_results}
\begin{adjustbox}{max width=\textwidth}
\begin{tabular}{@{}lcccc@{}}
\toprule
& \multicolumn{2}{c}{In-distribution}
& \multicolumn{2}{c}{Out-of-distribution} \\
\cmidrule(lr){2-3}
\cmidrule(lr){4-5}
Method & Turn 4 & Turn 8 & Turn 4 & Turn 8 \\
\midrule
Qwen2.5-7B-Instruct
& $10.4 \pm 0.8$ & $12.4 \pm 0.8$
& $5.6 \pm 0.6$ & $7.3 \pm 1.2$ \\
\quad + SFT
& $36.1 \pm 0.6$ & $47.6 \pm 2.4$
& $22.5 \pm 1.9$ & $33.9 \pm 1.4$ \\
\quad + Trajectory-level GRPO
& $44.2 \pm 1.6$ & $48.6 \pm 2.1$
& $49.2 \pm 2.2$ & $55.4 \pm 2.4$ \\
1.6T DeepSeek-V4-Pro \textsc{Think}~\citep{deepseekv4_report,deepseek_v4_pro}
& $53.6 \pm 1.8$ & $\mathbf{67.9 \pm 0.5}$
& $43.0 \pm 3.8$ & $58.7 \pm 0.6$ \\
\midrule
SLIC-\textsc{Greedy}
& $\mathbf{60.7 \pm 1.3}$ & $63.4 \pm 1.6$
& $54.2 \pm 1.3$ & $58.7 \pm 2.1$ \\
SLIC-\textsc{Random}
& $\mathbf{60.7 \pm 2.7}$ & $65.0 \pm 1.8$
& $50.0 \pm 5.3$ & $52.6 \pm 5.5$ \\
SLIC-\textsc{Softmax}
& $58.8 \pm 2.9$ & $64.6 \pm 1.5$
& $\mathbf{59.2 \pm 2.1}$ & $\mathbf{66.1 \pm 2.5}$ \\
\bottomrule
\end{tabular}
\end{adjustbox}
\end{table*}

Table~\ref{tab:original_results} shows that all three SLIC
variants outperform SFT on both splits at both horizons.
At four turns, Greedy and Random achieve the highest
in-distribution success, while Softmax performs best on
held-out topologies.
Softmax retains this OOD advantage at eight turns,
reaching 66.1\% Pass@1.
All SLIC variants improve beyond their four-turn training
horizon, demonstrating that the learned policies can
benefit from additional simulator interactions.

\subsection{Setting at 65~nm}
\label{app:qwen25_65nm}

We train Qwen2.5-7B-Instruct on approximately 2k tasks
at the 65\,nm PTM node, using Cadence Spectre for simulation.
RL uses SLIC-\textsc{Softmax} with group size $G=4$
for 800 updates, without the success-conditional slack bonus.
The joint-normalization run is used in the main
backbone comparison.
Separate normalization improves Pass@1 by
9.6 percentage points in-distribution and
3.8 points out-of-distribution
(Table~\ref{tab:qwen25_65nm}), supporting its
effectiveness on Qwen2.5-7B-Instruct.

\begin{table}[t]
\centering
\small
\setlength{\tabcolsep}{6pt}
\renewcommand{\arraystretch}{1.08}
\caption{
Qwen2.5-7B-Instruct four-specification Pass@1
(\%) at Turn~8.
}
\label{tab:qwen25_65nm}
\begin{tabular}{@{}lcc@{}}
\toprule
Normalization & In-distribution & Out-of-distribution \\
\midrule
Joint
& $32.2 \pm 0.8$ & $24.9 \pm 1.7$ \\
Separate
& $41.8 \pm 0.5$ & $28.7 \pm 1.1$ \\
\bottomrule
\end{tabular}
\end{table}
\section{Additional Experiments and Analysis}

\subsection{Training Dynamics}
\label{app:training_dynamics}

\begin{figure*}[t]
\centering
\includegraphics[width=\textwidth]{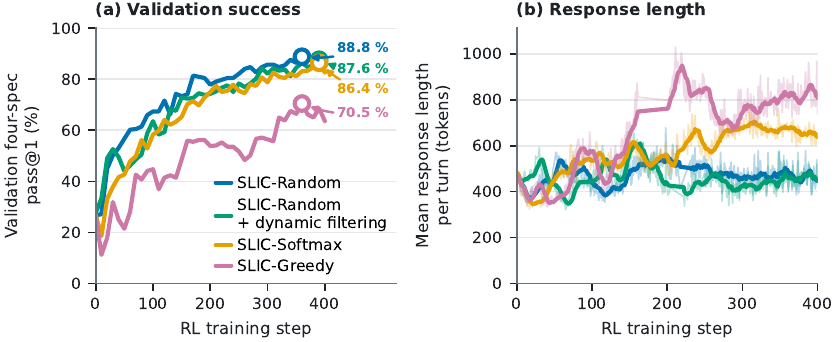}
\caption{
Training dynamics of SLIC under different continuation
strategies.
(a) Four-specification Pass@1 on the validation set used
for checkpoint selection; open circles mark the best
validation checkpoint for each variant.
(b) Mean generated response length per turn during
training, including both reasoning and actions.
}
\label{fig:training_dynamics}
\end{figure*}

Figure~\ref{fig:training_dynamics} shows that Greedy develops
the longest responses while remaining less
successful than Random and Softmax. Its sampled responses repeatedly reconsider edits, using
expressions such as ``Actually,'' ``Maybe,'' and ``let's revert''
(Figure~\ref{fig:reasoning_examples}).
This pattern is consistent with unproductive overthinking,
where additional tokens are spent revisiting decisions without
comparable performance gains~\citep{chen2025overthinking}.
Individual reflection markers are not sufficient evidence
of poor reasoning, as their effects depend on
context~\citep{jiang2026pathcal}.

Softmax's responses more often organize simulator feedback
into a diagnosis, proposed edits, and expected effects.
An auxiliary analysis of eight-turn evaluation rollouts also finds that its mean next-turn
improvement after a performance decrease remains similar
across the two splits, whereas Random's declines on held-out
topologies.
These observations suggest that Softmax uses additional
reasoning to support corrective progress on unfamiliar circuits.
Although longer training-time reasoning can benefit OOD
generalization~\citep{xue2026beyond}, our results indicate
that response length alone does not explain the differences
between continuation strategies.

\begin{figure*}[t]
\centering
\tcbset{
  sliccompact/.style={
    enhanced,
    width=\linewidth,
    equal height group=slicreasoningexamples,
    valign=top,
    boxrule=0.5pt,
    arc=1mm,
    boxsep=0pt,
    left=2mm,
    right=2mm,
    top=2mm,
    bottom=2mm,
    toptitle=1mm,
    bottomtitle=1mm,
    fonttitle=\small\bfseries,
    fontupper=\footnotesize,
    coltitle=black,
    before skip=0pt,
    after skip=0pt,
    before upper={
      \raggedright
      \setlength{\parindent}{0pt}
      \setlength{\parskip}{4pt}
    }
  }
}

\noindent
\begin{minipage}[t]{0.49\textwidth}
\vspace{0pt}
\begin{tcolorbox}[
  sliccompact,
  title={SLIC-\textsc{Softmax}},
  colback=blue!3,
  colframe=blue!45!black,
  colbacktitle=blue!12
]
[\ldots] \textbf{Diagnosis:} [\ldots]

3. Gain is adequate at 64 dB but will decrease after these
changes, which is acceptable since target is
$\geq 49.8$ dB.

\textbf{Strategy:} [\ldots]

- Increase \texttt{w\_MM12} slightly to increase output nMOS
current (increasing gm). [\ldots]

\textbf{Expected results after actions:}

- Bandwidth should increase from 11 kHz to $>75$ kHz

- Phase margin should improve from $-9^\circ$ to
$30$--$90^\circ$ [\ldots]
\end{tcolorbox}
\end{minipage}\hfill%
\begin{minipage}[t]{0.49\textwidth}
\vspace{0pt}
\begin{tcolorbox}[
  sliccompact,
  title={SLIC-\textsc{Greedy}},
  colback=orange!4,
  colframe=orange!55!black,
  colbacktitle=orange!15
]
[\ldots] Increasing gm or reducing parasitic?
\textbf{Actually} previous step moved in wrong direction.
[\ldots]

Let's \textbf{revert} a few to move back to PM 2 deg,
then try to push PM. [\ldots]

Let's \textbf{revert} some and also try to push gain.

\textbf{Maybe} increase second stage gain devices or first
stage. [\ldots]

Ok, let's try \textbf{revert} and try different.
\end{tcolorbox}
\end{minipage}

\caption{
Illustrative response excerpts from two different training
tasks.
Softmax organizes its response around diagnosis, proposed
edits, and expected effects.
Greedy repeatedly revisits the decision to revert earlier changes.
Bold text highlights these patterns, and brackets mark omissions.
}
\label{fig:reasoning_examples}
\end{figure*}

\subsection{Evaluation on Harder Tasks}
\label{app:hard_eval}

We additionally evaluate \textsc{Hard}, a more challenging
benchmark variant at the same 65~nm technology node, retaining
the 35 training topologies, six held-out topologies, and
parameter ranges of \textsc{Standard}
(Table~\ref{tab:standard_hard_stats}).
The fraction of valid initial designs decreases to 50.0\%
and 52.5\% on the in-distribution and out-of-distribution
evaluation splits, respectively.
To assess target difficulty, we use the dataset builder's
joint target-hit rate: the fraction of sampled candidate
designs satisfying all target specifications.
On held-out topologies, this rate drops from 2.3--3.4\%
on \textsc{Standard} to 0.18--0.40\% on \textsc{Hard},
indicating rarer joint target attainment
under the same sampling procedure.
\begin{table*}[t]
\centering
\small
\setlength{\tabcolsep}{5pt}
\renewcommand{\arraystretch}{1.08}
\caption{
Four-specification Pass@1 (\%) within eight turns on
\textsc{Hard}, evaluated on 420 in-distribution and
240 out-of-distribution tasks.
Reported overall and separately
for tasks with valid and invalid initial designs.
The upper group evaluates models trained on \textsc{Standard}
without additional fine-tuning; the lower group compares
training on \textsc{Hard} with and without the
gain-prioritized difficulty gate.
All models use Qwen3.5-4B unless otherwise specified.
}
\label{tab:hard_evaluation}
\begin{adjustbox}{max width=\textwidth}
\begin{tabular}{@{}lcccccc@{}}
\toprule
& \multicolumn{3}{c}{In-distribution}
& \multicolumn{3}{c}{Out-of-distribution} \\
\cmidrule(lr){2-4}
\cmidrule(lr){5-7}
Method
& Overall & Valid init. & Invalid init.
& Overall & Valid init. & Invalid init. \\
\midrule
\multicolumn{7}{@{}l}{
\textit{Trained on \textsc{Standard}; evaluated on \textsc{Hard}}
} \\
SLIC-\textsc{Greedy}
& $36.2 \pm 2.3$ & $36.7 \pm 2.5$ & $35.7 \pm 2.1$
& $32.5 \pm 2.9$ & $29.9 \pm 3.6$ & $35.4 \pm 3.5$ \\
SLIC-\textsc{Softmax}
& $54.3 \pm 0.4$ & $52.2 \pm 1.0$ & $56.3 \pm 0.7$
& $54.4 \pm 2.3$ & $51.9 \pm 2.6$ & $57.3 \pm 2.0$ \\
SLIC-\textsc{Random}
& $57.9 \pm 1.8$ & $56.2 \pm 1.3$ & $59.5 \pm 2.4$
& $53.9 \pm 0.5$ & $54.2 \pm 1.7$ & $53.5 \pm 2.3$ \\
\quad + dynamic filtering
& $59.9 \pm 1.6$ & $59.4 \pm 1.1$ & $60.5 \pm 2.1$
& $59.4 \pm 2.1$ & $56.1 \pm 2.8$ & $63.2 \pm 3.0$ \\
\quad w/o domain-validity gates
& $56.7 \pm 1.0$ & $55.6 \pm 0.7$ & $57.8 \pm 1.5$
& $49.2 \pm 0.7$ & $49.2 \pm 0.8$ & $49.1 \pm 1.8$ \\
Qwen3.5-9B SLIC-\textsc{Random}
& $62.5 \pm 1.6$ & $60.8 \pm 3.8$ & $64.3 \pm 0.8$
& $58.1 \pm 1.3$ & $58.2 \pm 0.5$ & $57.9 \pm 2.3$ \\
\midrule
\multicolumn{7}{@{}l}{
\textit{SLIC-\textsc{Random} trained and evaluated on \textsc{Hard}}
} \\
Without difficulty gate
& $58.0 \pm 0.5$ & $56.3 \pm 1.2$ & $59.7 \pm 2.0$
& $50.3 \pm 1.0$ & $48.9 \pm 3.7$ & $51.8 \pm 6.1$ \\
With difficulty gate
& $63.5 \pm 0.6$ & $60.8 \pm 1.5$ & $66.2 \pm 1.4$
& $56.5 \pm 1.0$ & $57.4 \pm 2.3$ & $55.6 \pm 1.0$ \\
\bottomrule
\end{tabular}
\end{adjustbox}
\end{table*}

\paragraph{Transfer from standard training.}
Table~\ref{tab:hard_evaluation} shows that policies trained on
\textsc{Standard} retain substantial success on \textsc{Hard},
with random and softmax continuation outperforming greedy
continuation on both splits.
The comparable success rates between initially valid and
invalid tasks may partly reflect our effort to balance
task difficulty during dataset construction.
Valid initial designs are selected to perform poorly on
the designated hard specification, while the selection
criteria tend to yield less demanding targets for invalid
initial designs, which additionally require recovery to
a valid operating state.

\paragraph{Gain-prioritized difficulty gate.}
This additional gate activates reward terms for easier
specifications only after gain meets its target within
the 5\% tolerance, preventing easier objectives
from dominating the reward before gain is sufficient
~\citep{liu2026gdpo}.
Specifically, the gain prerequisite is
$\Gamma_t \geq (1-0.05)\Gamma^\star$.
We compare training with and without this difficulty gate,
while retaining the domain-validity gates and the same
5\% success tolerance in both configurations.
Although we observed no clear benefit on \textsc{Standard},
enabling it during training on \textsc{Hard} improves
in-distribution and out-of-distribution Pass@1 by 5.5 and
6.2 percentage points, respectively
(Table~\ref{tab:hard_evaluation}).
The larger initial gain deficits on \textsc{Hard} suggest
that gain is a more prominent bottleneck, making gain
prioritization useful for identifying suitable operating points.
On \textsc{Standard}, the difficulty differences among
specifications are less pronounced, which may explain
the limited benefit of imposing this priority.

\subsection{Factors Associated with Topology Generalization}
\label{app:topology_factors}

\begin{figure*}[p]
\centering
\includegraphics[width=\linewidth]
{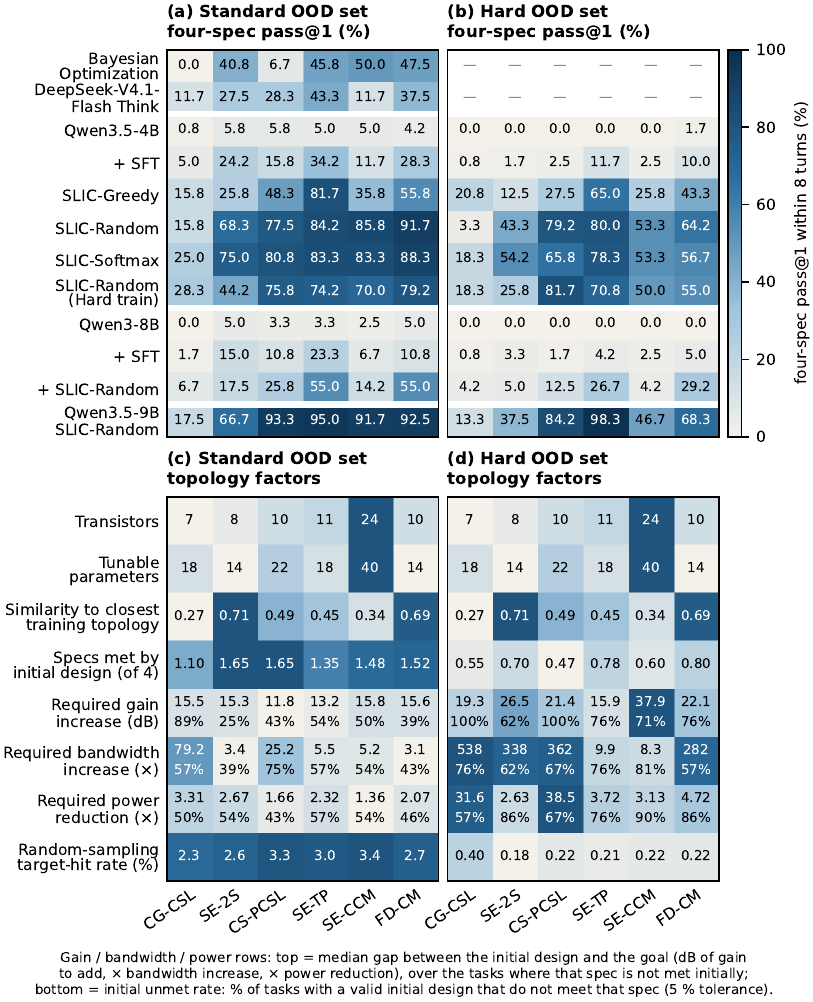}
\caption{
Topology-level performance and task characteristics on
\textsc{Standard} (left) and \textsc{Hard} (right).
(a,b) Four-specification Pass@1 (\%) on each held-out topology.
Language models use up to eight turns, and BO uses 192 simulator
calls per topology. SLIC checkpoints are trained on
\textsc{Standard} unless labeled otherwise.
Dashes denote unavailable results.
(c,d) Circuit characteristics and task difficulty.
For each specification gap, the main value is the median gap
among tasks that initially miss the target. The smaller
percentage gives the fraction of initially valid designs
that miss it. Color scales are shared across datasets:
one for Pass@1 and one per factor, with darker cells
indicating larger values.
Topology abbreviations:
CG-CSL, common-gate amplifier with a current-source load;
SE-2S, single-ended two-stage amplifier;
CS-PCSL, common-source amplifier with a PMOS cascode
current-source load;
SE-TP, single-ended telescopic PMOS amplifier;
SE-CCM, single-ended cascode current-mirror amplifier;
FD-CM, fully differential current-mirror amplifier.
}
\label{fig:ood_generalization_factors}
\end{figure*}

Figure~\ref{fig:ood_generalization_factors} highlights
contrasting cases of topology generalization.
On \textsc{Standard}, CG-CSL has the fewest transistors
among the held-out topologies, yet SLIC-\textsc{Random}
achieves only 15.8\% Pass@1.
Its tasks require large bandwidth increases and power
reductions, making this structurally simple circuit
challenging to optimize.
Conversely, SE-CCM has the most tunable parameters (40)
but reaches 85.8\% Pass@1, with smaller required bandwidth
increases and power reductions.
For the same SE-CCM topology, success drops to 53.3\%
on \textsc{Hard}, where the required gain improvement
is substantially larger.
These cases illustrate that circuit size alone does not
determine optimization difficulty: the performance changes
required by the tasks are also important for interpreting
success across topologies.

\begin{figure*}[t]
    \centering
    \includegraphics[width=\textwidth]{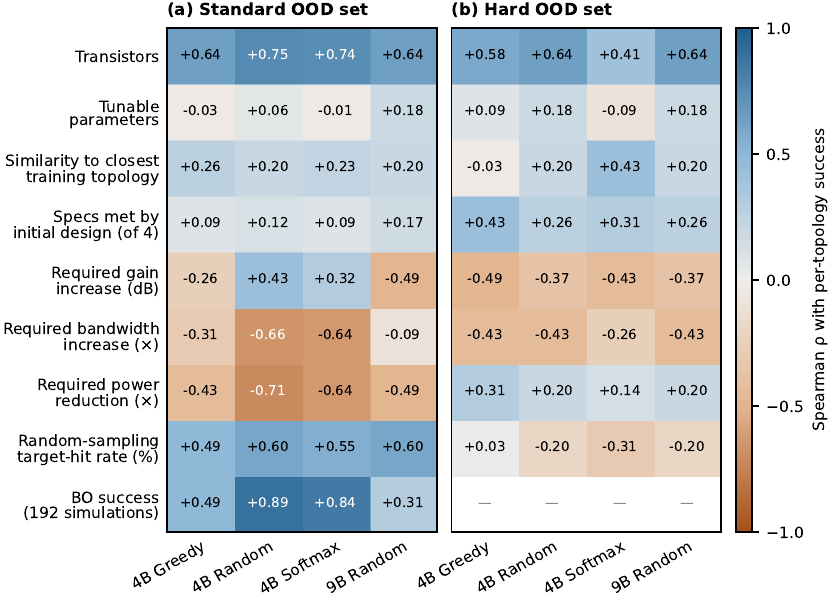}
    \caption{
Spearman rank correlations between topology-level factors
and mean four-specification Pass@1 within eight turns on
\textsc{Standard} (left) and \textsc{Hard} (right).
Each coefficient is computed across the six held-out topologies
for the indicated SLIC variant.
Gain, bandwidth, and power factors describe required
improvements rather than raw performance values;
larger values denote larger required changes.
Blue and orange indicate positive and negative associations
with success, respectively, on a shared scale centered at zero.
Dashes denote unavailable results.
}
\label{fig:topology_factor_correlations}
\end{figure*}

\paragraph{Correlation analysis.}
Figure~\ref{fig:topology_factor_correlations} relates
these performance differences to topology-level factors.
On \textsc{Standard}, larger required bandwidth increases
and power reductions are associated with lower success
for the 4B \textsc{Random} and \textsc{Softmax} variants
(Spearman $\rho=-0.71$ to $-0.64$).
On \textsc{Hard}, required gain increases become more
variable across topologies and are negatively associated
with success across all four variants, while the negative
association with required power reduction does not persist.
Transistor count is positively associated with success
on both datasets, whereas tunable parameter count shows
little association, indicating that larger circuits are
not necessarily harder to optimize in this benchmark.
This is consistent with our goal construction, which aims
to balance task difficulty across topologies:
even structurally simple circuits can be assigned demanding
performance targets and consequently exhibit low success rates.
Together, these observations support interpreting difficulty
through the performance changes required of each topology,
rather than circuit size alone.

\subsection{Distribution Shift}
\label{app:performance_space_shift}

\begin{figure*}[p]
\centering
\includegraphics[width=\textwidth]
{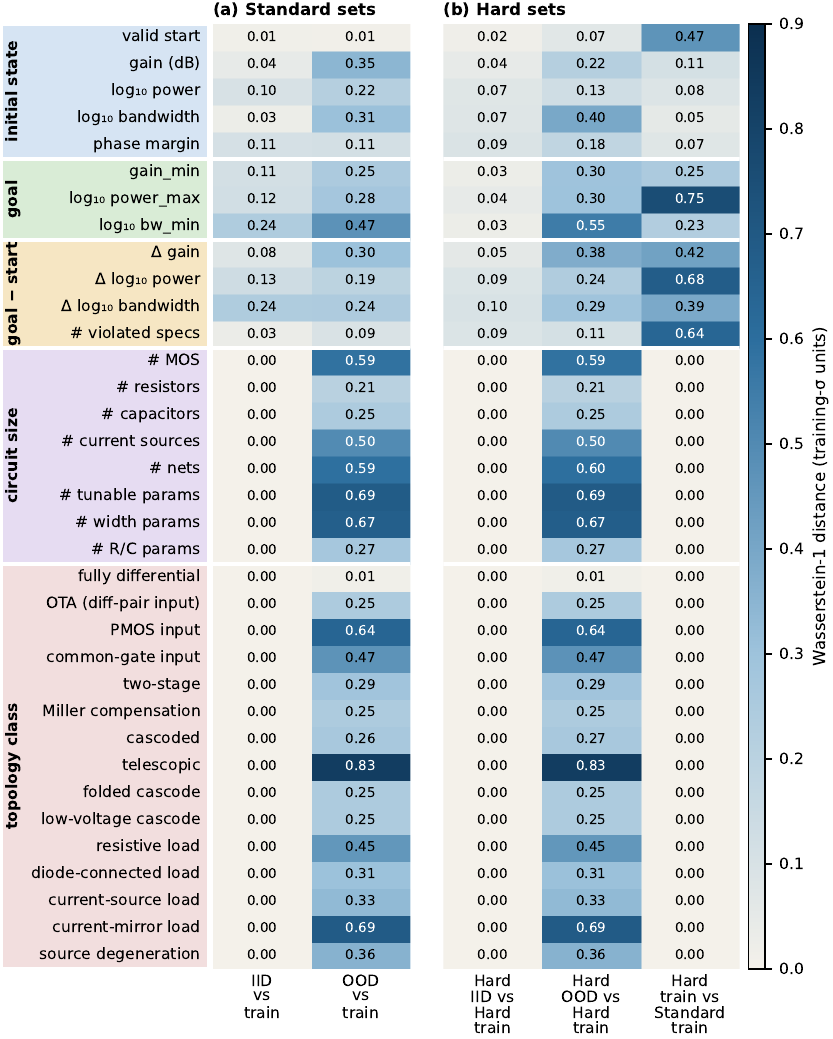}
\caption{
Feature-wise distribution shifts for \textsc{Standard}
and \textsc{Hard}.
Each column compares the indicated dataset with its reference
training set using Wasserstein-1 distances computed after standardization
using the same reference-training statistics for both
compared sets.
Power and bandwidth are log-transformed, and distances
are expressed in reference-training standard-deviation units.
Features describe initial conditions, specification targets,
initial target gaps, circuit-size counts, and 15 expert-defined
binary design descriptors.
The descriptors encode architectural properties using fixed
rules applied to circuit metadata and canonical topology names;
multiple descriptors may apply to the same topology.
For binary descriptors, distances reflect differences in
attribute prevalence after normalization.
Larger distances indicate stronger distribution shifts.
}
\label{fig:dataset_shift}
\end{figure*}

Figure~\ref{fig:dataset_shift} compares initial conditions,
target specifications, initial target gaps, and circuit
structure across dataset splits.
Within both \textsc{Standard} and \textsc{Hard},
in-distribution tasks closely match their respective
training sets, whereas held-out tasks show larger differences
in device and parameter counts and architectural properties.
Target specifications also differ across splits,
particularly for bandwidth, so held-out evaluation involves
changes in both circuit structure and target distributions.
Comparing \textsc{Hard} training data with \textsc{Standard},
the structural distributions remain unchanged, while
initial validity, target specifications, and initial target
gaps shift substantially, especially for power.
These changes reflect how \textsc{Hard} modifies the starting
conditions and performance requirements within the same
topology pool.

\subsection{Statistical Analysis}
\label{app:statistical_analysis}

We conduct task-paired comparisons on the 840
in-distribution and 240 out-of-distribution evaluation tasks.
Each task's score is the mean of three binary indicators
of four-specification success within eight turns.
We estimate mean task-wise differences and their 95\%
percentile confidence intervals using paired bootstrap
resampling over tasks (10{,}000 resamples, seed~7).
Each resampled task retains all three evaluation repetitions
and the pairing between methods.

Table~\ref{tab:key-significance} reports selected comparisons.
Significance is assessed using two-sided bootstrap tests
with Holm correction across all 20 Turn~8 tests in the analysis.
The analysis treats tasks as independent sampling units
and conditions on the evaluated checkpoints.

\begin{table*}[t]
\centering
\small
\setlength{\tabcolsep}{6pt}
\renewcommand{\arraystretch}{1.12}
\caption{
Task-paired differences in four-specification Pass@1
within eight turns.
Differences ($\Delta$, in percentage points) are computed
as the first method minus the second.
Random, Softmax, and Greedy denote the corresponding
SLIC variants.
The normalization and slack-bonus comparisons use their
respective matched training settings.
Confidence intervals are pointwise 95\% percentile
bootstrap intervals.
$^{*}$ indicates significance at $\alpha=0.05$ after
Holm correction across all 18 Turn~8 tests in the analysis.
}
\label{tab:key-significance}
\begin{adjustbox}{max width=\textwidth}
\begin{tabular}{@{}lcccc@{}}
\toprule
& \multicolumn{2}{c}{In-distribution}
& \multicolumn{2}{c}{Out-of-distribution} \\
\cmidrule(lr){2-3}
\cmidrule(lr){4-5}
Comparison
& $\Delta$ & 95\% CI
& $\Delta$ & 95\% CI \\
\midrule

Random vs.\ SFT
& $+60.6^{*}$ & $[57.9,\,63.3]$
& $+50.7^{*}$ & $[45.0,\,56.4]$ \\

Random vs.\ DeepSeek-V4.1-Flash \textsc{Think}
& $+48.5^{*}$ & $[45.6,\,51.5]$
& $+43.9^{*}$ & $[38.2,\,49.4]$ \\

Random vs.\ Greedy
& $+20.7^{*}$ & $[18.1,\,23.3]$
& $+26.7^{*}$ & $[21.2,\,32.1]$ \\

Random vs.\ w/o domain-validity gates
& $+4.3^{*}$ & $[2.4,\,6.2]$
& $+6.9^{*}$ & $[3.1,\,11.0]$ \\

Softmax vs.\ Random
& $-2.8^{*}$ & $[-4.7,\,-1.0]$
& $+2.1$ & $[-1.7,\,5.8]$ \\

\midrule
Separate vs.\ joint normalization
& $+6.2^{*}$ & $[4.2,\,8.3]$
& $+4.9$ & $[0.1,\,9.7]$ \\

With vs.\ without success-conditional slack bonus
& $+8.5^{*}$ & $[6.3,\,10.8]$
& $+12.1^{*}$ & $[7.5,\,16.7]$ \\

\bottomrule
\end{tabular}
\end{adjustbox}
\end{table*}

SLIC-\textsc{Random} significantly outperforms SFT, DeepSeek, SLIC-\textsc{Greedy}, and its variant
without domain-validity gates on both splits after correction.
SLIC-\textsc{Softmax} has significantly lower
in-distribution performance than SLIC-\textsc{Random},
while its higher out-of-distribution mean does not
establish a statistically significant difference. Among the component ablations, separate normalization
significantly improves in-distribution performance,
while its out-of-distribution improvement does not remain
significant after correction.
The success-conditional slack bonus significantly improves
performance on both splits under dynamic filtering.

\section{Implementation Details}
\label{app:config}

\subsection{Dataset}
\label{app:dataset}
We build our benchmark by adapting circuit topologies from the GANA dataset~\citep{kunal2020gana}. For each topology, we retain the main amplifier circuit together with its associated current-bias circuit, and curate the resulting circuits into train, validation, and held-out test splits for simulator-interactive optimization. We group the topologies by complexity, treating differential-input amplifiers as complex circuits and single-input amplifiers as simple circuits. The dataset is constructed by varying both the initial circuit states and the target goal specifications. Within the prescribed topology split, training and
evaluation tasks are sampled with stratification by
circuit topology and task difficulty to control the
representation of different circuit structures and
optimization demands. The resulting category-level dataset statistics are summarized in Table~\ref{tab:standard_hard_stats}.

\paragraph{SFT data.}
The SFT corpus contains 6{,}981 single-turn samples
from 1{,}568 teacher-generated episodes across the
35 training topologies.
We split the corpus by episode into 6{,}233 training
samples from 1{,}400 episodes and 748 test samples
from 168 episodes.
All samples from the same episode remain in one split,
while both splits cover the same 35 topologies.

\begin{table*}[h]
\centering
\small
\setlength{\tabcolsep}{5pt}
\renewcommand{\arraystretch}{1.15}
\caption{
Category-wise dataset statistics for \textsc{Standard}
and \textsc{Hard} at the 65~nm technology node.
Both variants share the topology sets and parameter ranges
within each split and category.
Tunable parameters are reported as min--max ranges across
topologies.
Valid init.\ denotes the percentage of tasks with a valid
initial design.
}
\label{tab:standard_hard_stats}
\begin{adjustbox}{max width=\textwidth}
\begin{tabular}{@{}cccccccc@{}}
\toprule
& & & &
\multicolumn{2}{c}{\textsc{Standard}} &
\multicolumn{2}{c}{\textsc{Hard}} \\
\cmidrule(lr){5-6}
\cmidrule(lr){7-8}
\textbf{Split}
& \textbf{Category}
& \shortstack{\textbf{Topology}\\\textbf{count}}
& \shortstack{\textbf{Tunable parameters}\\\textbf{(min--max)}}
& \textbf{Samples}
& \shortstack{\textbf{Valid init.}\\\textbf{(\%)}}
& \textbf{Samples}
& \shortstack{\textbf{Valid init.}\\\textbf{(\%)}} \\
\midrule

\multirow{3}{*}{Train}
& Complex & 19 & 10--36
& 3,605 & 70.5 & 2,432 & 49.2 \\
& Simple & 16 & 4--22
& 3,027 & 70.4 & 2,048 & 49.2 \\
& \textbf{Total} & 35 & 4--36
& 6,632 & 70.4 & 4,480 & 49.2 \\

\midrule

\multirow{3}{*}{\shortstack{In-distribution\\test}}
& Complex & 19 & 10--36
& 456 & 71.1 & 228 & 50.0 \\
& Simple & 16 & 4--22
& 384 & 70.8 & 192 & 50.0 \\
& \textbf{Total} & 35 & 4--36
& 840 & 71.0 & 420 & 50.0 \\

\midrule

\multirow{3}{*}{\shortstack{Out-of-distribution\\test}}
& Complex & 4 & 14--40
& 160 & 70.0 & 160 & 52.5 \\
& Simple & 2 & 18--22
& 80 & 70.0 & 80 & 52.5 \\
& \textbf{Total} & 6 & 14--40
& 240 & 70.0 & 240 & 52.5 \\

\bottomrule
\end{tabular}
\end{adjustbox}
\end{table*}

\subsection{Hosted Model Evaluation}
\label{app:hosted_model_eval}

We evaluated DeepSeek-V4.1-Flash on the 840 ID tasks
in two 420-task batches at different times, using the
same API model identifier, prompts, and protocol.
DeepSeek achieved 35.3\% and 41.9\% Pass@1 on the two
batches, for an aggregate of 38.6\% in Table~\ref{tab:val_turns}. GPT-6-Astra scored similarly on the two batches
(91.0\% and 90.2\%), suggesting that the gap for DeepSeek
may reflect changes in the served model or inference backend.
Using fixed local checkpoints avoids uncertainty about
model version changes between evaluations.
Across both splits, we spent about \$227 in API charges
to evaluate DeepSeek three times per task and about \$276
to evaluate GPT-6-Astra once per task.

\subsection{Training Configurations}
\begin{table}[H]
\centering
\small
\caption{Main supervised fine-tuning (SFT) configuration.}
\label{tab:sft_config}
\begin{tabular}{lll}
\toprule
\textbf{Category} & \textbf{Setting} & \textbf{Value} \\
\midrule

Hardware
& Nodes / GPUs & 1 node, 8 $\times$ H100 (80 GB) \\
\midrule

Model
& Base model & Qwen3.5-4B \\
& Training strategy & FSDP2 \\
& Model load dtype & bfloat16 \\
& Gradient checkpointing & Enabled \\
\midrule

Data
& Dataset & Single-turn step samples \\
& Training samples & 6{,}233 \\
& Test samples & 748 \\
& Train batch size & 8 \\
& Micro-batch size / GPU & 1 \\
& Max sequence length & 11,776 \\
& Truncation & Left \\
\midrule

Environment / Task
& Representation & XML \\
& Technology node & 65~nm (PTM) \\
& Max episode steps & 8 \\
& History length & 1 \\
\midrule

Optimization
& Optimizer & AdamW \\
& Learning rate & $1\times10^{-5}$ \\
& Warmup ratio & 0.1 \\
& Total epochs & 4 \\
\bottomrule
\end{tabular}
\end{table}

\begin{table}[htbp]
\centering
\small
\caption{Main RL training configuration for SLIC-\textsc{Random}.}
\label{tab:rl_config}
\begin{adjustbox}{max width=\linewidth}
\begin{tabular}{lll}
\toprule
\textbf{Category} & \textbf{Setting} & \textbf{Value} \\
\midrule

Hardware
& Nodes / GPUs & 1 node, 2 $\times$ B200 \\
& CPUs per GPU & 16 \\
& Host memory & 400 GB \\
\midrule

Model
& Initialization checkpoint & Qwen3.5-4B SFT \\
& Actor learning rate & $1\times10^{-6}$ \\
& Learning-rate schedule & Cosine decay to $1\times10^{-7}$ \\
& Warmup & None \\
& Optimizer & AdamW \\
& Weight decay & 0.01 \\
& Gradient checkpointing & Enabled \\
\midrule

Data
& Train batch size & 16 tasks \\
& Rollout group size & 8 \\
& Training turns per step & 1,024 \\
& Validation set size & 420 tasks \\
& Max prompt length & 13,072 \\
& Max response length & 2,048 \\
& Truncation & Middle \\
\midrule

Environment
& Representation & XML \\
& Technology node & 65~nm (PTM) \\
& Max steps & 8 \\
& History length & 1 \\
\midrule

RL algorithm
& Algorithm & GRPO with RGRS \\
& Advantage normalization & Separate validity / performance \\
& Validity / performance weights & 1.0 / 1.0 \\
& PPO epochs & 1 \\
& PPO mini-batch size & 256 \\
& PPO micro-batch size / GPU & 1 \\
& Clip ratio (low / high) & 0.2 / 0.28 \\
& KL loss & Disabled \\
\midrule

Rollout
& Backend & vLLM \\
& Tensor parallel size & 1 \\
& Temperature & 1.0 \\
& Top-$p$ & 0.9 \\
& Samples per prompt & 1 \\
\midrule

Reward shaping
& Invalid action penalty magnitude & 0.1 \\
& Simulation failure penalty magnitude & 0.1 \\
\midrule

Training schedule
& Total training steps & 400 \\
\bottomrule
\end{tabular}
\end{adjustbox}
\end{table}

\section{Limitations}
\label{app:limitations}
While SLIC demonstrates strong capabilities in analog circuit optimization, several limitations remain. First, SLIC focuses on device sizing within a given
circuit topology.
Extending it to a complete design flow that integrates
topology generation, sizing, robustness verification
under process, voltage, and temperature (PVT) variations,
and physical layout remains future work. Second, our evaluation currently focuses on amplifier
optimization with four specifications.
Extending SLIC to a broader range of circuits, including
low-dropout regulators (LDOs), low-noise amplifiers (LNAs),
and power amplifiers (PAs), and incorporating additional
coupled objectives such as noise and offset remain
directions for future work.

\section{Broader Impact}
\label{app:impact}

SLIC aims to advance practical AI-driven automation for analog circuit design, a field that remains highly manual, expert-intensive, and difficult to scale across circuit families, specifications, and technology nodes. By training a locally deployable language-model agent to iteratively optimize circuits using verifiable simulator feedback, this work points toward design tools that can reduce repetitive manual tuning and improve the scalability of analog and mixed-signal development. The local deployment setting is especially important for industry, where proprietary PDKs, device models, and circuit IP often cannot be exposed to hosted LLM APIs. Circuit-specialized small models could therefore provide a practical alternative for secure in-house design assistance and parallel multi-agent workflows across circuit sub-blocks. 

Local deployment could also support education in
analog integrated circuit design, particularly when
repeated use of hosted LLM APIs is costly and
confidentiality restrictions prevent sharing proprietary
commercial PDK data with external services.
Teaching analog integrated circuit design often requires
instructors to demonstrate how to make design decisions,
interpret simulation results, and debug circuits,
yet such individual guidance can be difficult to provide
when teaching resources are limited.
Analog integrated circuit design agents running locally
on students' computers could complement this guidance
by making the design process visible and interactive.
Students could follow proposed changes, inspect
simulator feedback, and explore subsequent revisions,
making practical design experience more accessible.

However, such systems should remain coupled with expert review and standard verification flows, since final analog designs must satisfy reliability, manufacturability, and safety requirements beyond the simulated objectives considered in this work.

\begingroup
\raggedright
\section{Case Studies}
\label{app:case}

We present two evaluation rollouts of SLIC-\textsc{Random}
at the 65~nm technology node, covering different topologies,
initial operating conditions, and target specifications.
Case~1 uses a fully-differential folded-cascode amplifier from
the in-distribution split and starts from a valid design that
meets only the bandwidth target.
Case~2 uses a held-out fully-differential current-mirror amplifier
from the out-of-distribution split and starts from an invalid design.
Both reach all four targets in four turns, including recovery
from an invalid intermediate design.
Model responses are reproduced as generated, including their
intermediate interpretations.
Long graph descriptions and repeated device features are omitted
where indicated.

\subsection{Case 1: In-Distribution Folded-Cascode Amplifier}
\label{app:case_study_1}

Figure~\ref{fig:case1_schematic} shows the folded-cascode
amplifier used in this case. This topology contains 20 transistors across the amplifier core
and three bias-generator blocks, with 31 tunable parameters.
The initial gain is $36.7$~dB below the target, power is about
$34$ times the budget, and phase margin is outside the required window.
The policy first recovers from an invalid update, then increases
bandwidth while retaining sufficient gain and acceptable power.

\begin{figure}[h]
\raggedright
\includegraphics[width=\linewidth]{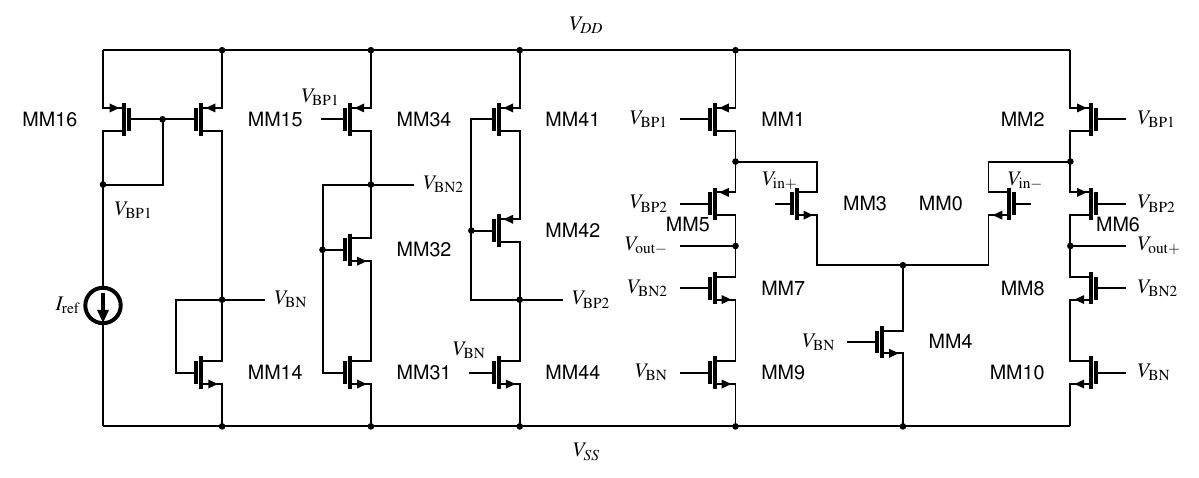}
\caption{Fully differential folded-cascode amplifier used in Case~1.
X0 is the amplifier core and X1--X3 generate its bias voltages.
Device identifiers and connections follow the supplied netlist.
In the top-level instance, $V_{\mathrm{biasn1}}$ is tied to
$V_{\mathrm{biasn}}$.
Identical bias labels denote the same electrical net.
}
\label{fig:case1_schematic}
\end{figure}

\begin{taskbox}{Task Setup}
\begin{tabularx}{\linewidth}{@{}l>{\raggedright\arraybackslash}X@{}}
\toprule
\textbf{Topology} & Fully-differential folded-cascode amplifier \\
\textbf{Evaluation split} & In-distribution \\
\textbf{Technology} & 65~nm CMOS (PTM), $V_{DD}=1.2$~V \\
\midrule
\textbf{Gain target} & $\geq 61.7$~dB \\
\textbf{Power target} & $\leq 425.7~\mu$W \\
\textbf{Bandwidth target} & $\geq 43.1$~kHz ($-3$~dB bandwidth) \\
\textbf{Phase-margin target} & $30^\circ \leq \mathrm{PM} \leq 90^\circ$ \\
\textbf{Evaluation tolerance} & $5\%$, as defined in the main evaluation \\
\midrule
\textbf{Initial gain} & $25.0$~dB \\
\textbf{Initial power} & $14{,}349~\mu$W \\
\textbf{Initial bandwidth} & $111$~kHz \\
\textbf{Initial phase margin} & $-36^\circ$ \\
\textbf{Initial in-band peaking} & $0.0$~dB \\
\textbf{Initial validity} & Valid. Only the bandwidth target is met. \\
\bottomrule
\end{tabularx}
\end{taskbox}

\begin{obsbox}{Structured System Input Excerpt}
\begin{verbatim}
<GRAPH_INFO>
TOPOLOGY_NAME: Fully_differential_folded_cascode
</GRAPH_INFO>
<GRAPH_NODE_IDMAP>
type=pin: @0=vdd @1=vss
type=current_source: @70=X1.I0
type=n_mosfet: @10=X0.MM0 @114=X3.MM44 @16=X0.MM10 @30=X0.MM3 @35=X0.MM4 @50=X0.MM7
  @55=X0.MM8 @60=X0.MM9 @73=X1.MM14 @88=X2.MM31 @93=X2.MM32
type=p_mosfet: @104=X3.MM41 @109=X3.MM42 @15=X0.MM1 @25=X0.MM2 @40=X0.MM5 @45=X0.MM6
  @78=X1.MM15 @83=X1.MM16 @98=X2.MM34
type=net: LG_Vbiasn LG_Vbiasn2 LG_Vbiasp1 LG_Vbiasp2 Vinn Vinp Voutn Voutp ...
</GRAPH_NODE_IDMAP>
<PINMAP> ... </PINMAP>
<EDGES> ... </EDGES>

Design bounds:
For tech 65nm, length l must stay within [45nm, 100um] and width w must stay
within [45nm, 500um]. Designs outside this range are invalid.
Fixed global: vdd=1.2V (user-set, not tunable).

Allowed valid target handles and parameter tokens:
  @10: l_MM30, w_MM30      (MM0 -- NMOS input pair, matched with @30)
  @15: l_MM12, w_MM12      (MM1 -- PMOS current source, matched with @25)
  @16: l_MM910, w_MM910    (MM10 -- NMOS current source, matched with @60)
  @35: l_MM4, w_MM4        (MM4 -- NMOS tail)
  @40: l_MM56, w_MM56      (MM5 -- PMOS cascode, matched with @45)
  @50: l_MM78, w_MM78      (MM7 -- NMOS cascode, matched with @55)
  @70: iref                (I0 -- reference current)
  @73, @78, @83: MM14/MM15/MM16   (X1: LG_nmos_l1 -> Vbiasn, Vbiasp1)
  @88, @93, @98: MM31/MM32/MM34   (X2: LG_load_biasn_LV -> Vbiasn2)
  @104, @109, @114: MM41/MM42/MM44 (X3: LG_load_biasp_LV -> Vbiasp2)
\end{verbatim}
\end{obsbox}

\begin{obsbox}{Turn 1 User Input Excerpt}
\begin{verbatim}
Your goal is: Achieve gain >= 61.7 dB, power <= 425.7 uW,
-3dB bandwidth >= 43.1 kHz,
30.0 deg <= phase_margin <= 90.0 deg
(each spec has 5% tolerance). Prioritize gain first.
[...]

Your current observation is:
<PERFORMANCE>
<GainAt1kHz>25.0 dB</GainAt1kHz>
<Power>14349 uW</Power>
<Bandwidth3dB>111 kHz</Bandwidth3dB>
<PhaseMargin>-36 deg</PhaseMargin>
<InBandFlatness>0.0 dB (limit 6)</InBandFlatness>
</PERFORMANCE>
<FEATURES>
@10 id=X0.MM0 type=n_mosfet w=w_MM30=34.78u l=l_MM30=166.8n Vgs=496 mV
    Vds=698 mV Vdsat=83.9 mV Ids=438 uA Gm=7 mS region=sat
@15 id=X0.MM1 type=p_mosfet w=w_MM12=17.5u l=l_MM12=69.58n Vgs=-476 mV
    Vds=-286 mV Vdsat=-157 mV Ids=-436 uA Gm=4.6 mS region=sat
@16 id=X0.MM10 type=n_mosfet w=w_MM910=175.1u l=l_MM910=6.297u Vgs=834 mV
    Vds=5 mV Vdsat=367 mV Ids=33.6 uA Gm=66.1 uS region=triode
@35 id=X0.MM4 type=n_mosfet w=w_MM4=4.998u l=l_MM4=199.9n Vgs=834 mV
    Vds=216 mV Vdsat=315 mV Ids=872 uA Gm=2.2 mS region=triode
@40 id=X0.MM5 type=p_mosfet w=w_MM56=117.9n l=l_MM56=20u Vgs=-914 mV
    Vds=-202 mV Vdsat=-471 mV Ids=-22.0 nA Gm=48.2 nS region=triode
@50 id=X0.MM7 type=n_mosfet w=w_MM78=75.97n l=l_MM78=20u Vgs=502 mV
    Vds=706 mV Vdsat=114 mV Ids=11.9 nA Gm=167 nS region=sat
@70 id=X1.I0 type=current_source dc=iref=23u ...
</FEATURES>
\end{verbatim}
\end{obsbox}
\begin{reasonbox}{Turn 1 Original LLM Output Excerpt}
\begin{verbatim}
<think>
The current operating point is far from specs: DC gain is only 25 dB (needs >=61.7 dB),
power is 14.35 mW (needs <=425.7 uW), and phase margin is -36 deg (needs 30-90 deg).
Bandwidth is 111 kHz, which is acceptable, but the main problems are: 1. Excessive bias
current: iref=23u causes excessive DC current. 2. PMOS current source MM1/MM2 is too
short (69.58nm) and wide, allowing too much DC current. 3. NMOS current source MM9/MM10
is very wide (175.1u) and acts as resistor rather than current source; needs moderate
W/L. 4. Input pair MM3/MM0 is too wide and short; needs stronger gm with better ro.
5. Tail MM4 is over-sized in current. 6. PMOS current source MM5/MM6 and NMOS cascode
   MM7/MM8 are weak (only 117.9n/75.97n). 7. DC gain is low because ro of signal devices
   are too small. Strategy: Reduce iref, make PMOS load stronger but less conductive,
   improve input pair, moderate tail, strengthen cascode current source, and fix bias
   blocks. ...
</think>
<action>decrease @70 iref 95%</action>
<action>increase @15 l_MM12 200%</action>
<action>decrease @15 w_MM12 50%</action>
<action>decrease @30 w_MM30 70%</action>
<action>increase @30 l_MM30 200%</action>
<action>decrease @35 w_MM4 50%</action>
<action>increase @35 l_MM4 20%</action>
<action>decrease @16 w_MM910 95%</action>
<action>increase @16 l_MM910 50%</action>
<action>increase @40 w_MM56 500%</action>
<action>decrease @40 l_MM56 99%</action>
<action>increase @50 w_MM78 500%</action>
<action>decrease @50 l_MM78 99%</action>
<action>decrease @73 w_MM14 95%</action>
<action>decrease @73 l_MM14 85%</action>
<action>decrease @78 w_MM15 98%</action>
<action>increase @78 l_MM15 300%</action>
<action>decrease @83 w_MM16 95%</action>
<action>decrease @83 l_MM16 70%</action>
<action>increase @104 w_MM41 500%</action>
<action>increase @109 w_MM42 500%</action>
<action>increase @114 w_MM44 10%</action>
\end{verbatim}
\end{reasonbox}
\begin{resultbox}{Simulator Result After Turn 1}
\begin{tabular}{@{}ll@{}}
\textbf{Gain / power} & $-21.9$~dB / $23~\mu$W \\
\textbf{Bandwidth} & Not measurable \\
\textbf{Phase margin} & Unavailable \\
\end{tabular}

\smallskip
The 22 simultaneous updates produce an invalid design with
non-positive gain. The policy receives the simulator measurements
and the \texttt{design\_error} message before the next turn.
\end{resultbox}
\begin{obsbox}{Turn 2 User Input Excerpt}
\begin{verbatim}
<PERFORMANCE>
<Status>INVALID - design_error</Status>
<Error>Design produces non-positive gain: -21.86 dB.</Error>
<GainAt1kHz>-21.9 dB</GainAt1kHz>
<Power>23 uW</Power>
<Bandwidth3dB>not measurable</Bandwidth3dB>
<Validity>non-positive gain; power/bandwidth/PM not meaningful until gain > 0</Validity>
</PERFORMANCE>
<FEATURES>
@10 id=X0.MM0 type=n_mosfet w=w_MM30=10.4u l=l_MM30=500.4n
@15 id=X0.MM1 type=p_mosfet w=w_MM12=8.8u l=l_MM12=208.7n
@16 id=X0.MM10 type=n_mosfet w=w_MM910=8.8u l=l_MM910=9.4u
@35 id=X0.MM4 type=n_mosfet w=w_MM4=2.5u l=l_MM4=239.9n
@40 id=X0.MM5 type=p_mosfet w=w_MM56=707.4n l=l_MM56=200n
@50 id=X0.MM7 type=n_mosfet w=w_MM78=455.8n l=l_MM78=200n
@70 id=X1.I0 type=current_source dc=iref=1.2u ...
</FEATURES>
\end{verbatim}
\end{obsbox}
\begin{reasonbox}{Turn 2 Original LLM Output Excerpt}
\begin{verbatim}
<think>
The current operating point is invalid: DC gain is -21.9 dB (magnitude only 0.219x). The
previous aggressive changes destroyed the DC bias: the PMOS current source/load
(MM1/MM2) is too strong, tail MM4 is weak, input pair MM3/MM0 is weak, and NMOS current
source MM9/MM10 is overkill. The core needs to be restored: make PMOS load weaker
(larger l_MM12, smaller MM1/MM2), stronger input pair MM3/MM0, stronger tail MM4,
moderate iref, and adjust NMOS cascode/load. Need positive DC gain first.
</think>
<action>increase @15 l_MM12 400.0%</action>
<action>decrease @15 w_MM12 30.0%</action>
<action>increase @10 w_MM30 100.0%</action>
<action>increase @10 l_MM30 100.0%</action>
<action>increase @35 w_MM4 300.0%</action>
<action>increase @35 l_MM4 50.0%</action>
<action>decrease @16 w_MM910 70.0%</action>
<action>increase @16 l_MM910 20.0%</action>
<action>increase @40 w_MM56 300.0%</action>
<action>increase @40 l_MM56 20.0%</action>
<action>increase @50 w_MM78 200.0%</action>
<action>increase @50 l_MM78 20.0%</action>
<action>decrease @70 iref 10.0%</action>
<action>decrease @73 w_MM14 70.0%</action>
<action>decrease @73 l_MM14 20.0%</action>
<action>increase @78 w_MM15 100.0%</action>
<action>decrease @78 l_MM15 10.0%</action>
<action>decrease @83 w_MM16 70.0%</action>
<action>increase @83 l_MM16 10.0%</action>
<action>increase @98 w_MM34 150.0%</action>
<action>decrease @93 w_MM32 85.0%</action>
<action>decrease @88 w_MM31 70.0%</action>
<action>decrease @88 l_MM31 60.0%</action>
\end{verbatim}
\end{reasonbox}
\begin{resultbox}{Simulator Result After Turn 2}
\begin{tabular}{@{}ll@{}}
\textbf{Gain / power} & $75.6$~dB / $98~\mu$W \\
\textbf{Bandwidth / phase margin} & $15$~kHz / $42^\circ$ \\
\textbf{In-band peaking} & $0.0$~dB \\
\end{tabular}

\smallskip
The design becomes valid and meets the gain, power, and
phase-margin targets. Bandwidth remains below its $43.1$~kHz target,
while gain now has substantial headroom.
\end{resultbox}
\begin{obsbox}{Turn 3 User Input Excerpt}
\begin{verbatim}
<PERFORMANCE>
<GainAt1kHz>75.6 dB</GainAt1kHz>
<Power>98 uW</Power>
<Bandwidth3dB>15 kHz</Bandwidth3dB>
<PhaseMargin>42 deg</PhaseMargin>
<InBandFlatness>0.0 dB (limit 6)</InBandFlatness>
</PERFORMANCE>
<FEATURES>
@10 id=X0.MM0 type=n_mosfet w=w_MM30=20.8u l=l_MM30=1u Vgs=324 mV
    Vds=434 mV Vdsat=48.5 mV Ids=3 uA Gm=63.6 uS region=subthreshold
@15 id=X0.MM1 type=p_mosfet w=w_MM12=6.2u l=l_MM12=1u Vgs=-477 mV
    Vds=-760 mV Vdsat=-139 mV Ids=-3.3 uA Gm=42.6 uS region=sat
@16 id=X0.MM10 type=n_mosfet w=w_MM910=2.6u l=l_MM910=11.3u Vgs=446 mV
    Vds=281 mV Vdsat=83.1 mV Ids=351 nA Gm=6 uS region=sat
@35 id=X0.MM4 type=n_mosfet w=w_MM4=10u l=l_MM4=359.8n Vgs=446 mV
    Vds=6.2 mV Vdsat=82.0 mV Ids=6 uA Gm=71.8 uS region=triode
@40 id=X0.MM5 type=p_mosfet w=w_MM56=2.8u l=l_MM56=240n Vgs=-440 mV
    Vds=-110 mV Vdsat=-61.9 mV Ids=-342 nA Gm=7.5 uS region=subthreshold
@50 id=X0.MM7 type=n_mosfet w=w_MM78=1.4u l=l_MM78=240n Vgs=345 mV
    Vds=49.2 mV Vdsat=45.8 mV Ids=341 nA Gm=7.7 uS region=cutoff
@70 id=X1.I0 type=current_source dc=iref=1.1u ...
</FEATURES>
\end{verbatim}
\end{obsbox}
\begin{reasonbox}{Turn 3 Original LLM Output Excerpt}
\begin{verbatim}
<think>
Current specs: Gain is 75.6 dB, well above the 61.7 dB target. Power is only 98 uW, well
within 425.7 uW. Phase margin is 42 deg, which meets 30-90 deg. The unmet spec is
bandwidth: only 15 kHz, needs >=43.1 kHz. The main bottleneck is the high output
resistance: MM3/MM0 are long (1u), MM1/MM2 PMOS load is long (1u), MM9/MM10 current
source is very long (11.3u), and PMOS/NMOS cascode/load devices MM5/MM6 and MM7/MM8 are
modest. The DC gain is excessive because ro of input pair and PMOS load are large. Need
to reduce device lengths, especially ro of signal path devices ... Goal is to lower ro
to reduce Rout, improving -3dB bandwidth while maintaining adequate DC gain.
</think>
<action>decrease @10 l_MM30 70%</action>
<action>increase @10 w_MM30 10%</action>
<action>decrease @15 l_MM12 60%</action>
<action>increase @15 w_MM12 20%</action>
<action>decrease @16 l_MM910 80%</action>
<action>increase @16 w_MM910 50%</action>
<action>increase @40 w_MM56 100%</action>
<action>decrease @40 l_MM56 20%</action>
<action>increase @50 w_MM78 100%</action>
<action>decrease @50 l_MM78 15%</action>
<action>decrease @70 iref 10%</action>
\end{verbatim}
\end{reasonbox}
\begin{resultbox}{Simulator Result After Turn 3}
\begin{tabular}{@{}ll@{}}
\textbf{Gain / power} & $75.9$~dB / $169~\mu$W \\
\textbf{Bandwidth / phase margin} & $37$~kHz / $52^\circ$ \\
\end{tabular}

\smallskip
Bandwidth increases from $15$ to $37$~kHz following the combined
length, width, and bias updates. It remains below the target even
with the $5\%$ evaluation tolerance. The other three targets remain met.
\end{resultbox}
\begin{obsbox}{Turn 4 Observation Summary}
\begin{verbatim}
(Observation: <GainAt1kHz>75.9 dB</GainAt1kHz><Power>169 uW</Power><Bandwidth3dB>37 kHz
 </Bandwidth3dB><PhaseMargin>52 deg</PhaseMargin>; input pair w=22.9u l=300n Ids=7.3 uA
 Gm=160 uS; MM1 w=7.4u l=400n; MM10 w=3.9u l=2.3u; MM5 w=5.6u l=192n; MM7 w=2.8u l=204n;
 iref=990n)
\end{verbatim}
\end{obsbox}
\begin{reasonbox}{Turn 4 Original LLM Output Excerpt}
\begin{verbatim}
<think>
Current specs: Gain is 75.9 dB (well above 61.7 dB target), Power is only 169 uW (well
within 425.7 uW), Phase Margin is 52 deg (meets 30-90 deg). The only unmet spec is
Bandwidth: 37 kHz requires >=43.1 kHz. The main bottleneck is the DC gain, dominated by
ro of signal path devices. Need to reduce ro of MM0/MM3 (input pair), MM1/MM2 PMOS load,
MM9/MM10 current source, MM5/MM6 PMOS cascode, MM7/MM8 NMOS cascode. Strategy: decrease
device lengths, especially l_MM30, l_MM12, l_MM910, l_MM56, l_MM78. Also improve iref
slightly for more current.
</think>
<action>decrease @10 l_MM30 50.0%</action>
<action>decrease @15 l_MM12 40.0%</action>
<action>increase @15 w_MM12 10.0%</action>
<action>decrease @16 l_MM910 55.0%</action>
<action>increase @16 w_MM910 15.0%</action>
<action>decrease @40 l_MM56 25.0%</action>
<action>increase @40 w_MM56 10.0%</action>
<action>decrease @50 l_MM78 30.0%</action>
<action>increase @50 w_MM78 15.0%</action>
<action>increase @70 iref 15.0%</action>
\end{verbatim}
\end{reasonbox}
\begin{resultbox}{Final Simulator Result After Turn 4}
\begin{tabular}{@{}ll@{}}
\textbf{Gain / power} & $68.4$~dB / $318~\mu$W \\
\textbf{Bandwidth / phase margin} & $172$~kHz / $56.4^\circ$ \\
\textbf{In-band peaking} & $0.0$~dB \\
\end{tabular}

\smallskip
All four original targets are satisfied without relying on
the evaluation tolerance. The episode terminates after four turns.
\end{resultbox}

\subsection{Case 2: Out-of-Distribution Current-Mirror Amplifier}
\label{app:case_study_2}

Figure~\ref{fig:case2_schematic} shows the current-mirror
amplifier used in this case. The second task starts with a gain of $-123.2$~dB and
$61.2$~dB of in-band peaking.
The input pair and PMOS output mirror are in triode,
and bandwidth is not measurable.
After two turns, the policy reaches a valid operating point
with sufficient gain and bandwidth.
Two further updates reduce power and improve phase margin
while retaining enough gain to meet the target.

\begin{figure}[H]
\raggedright
\includegraphics[width=0.92\linewidth]{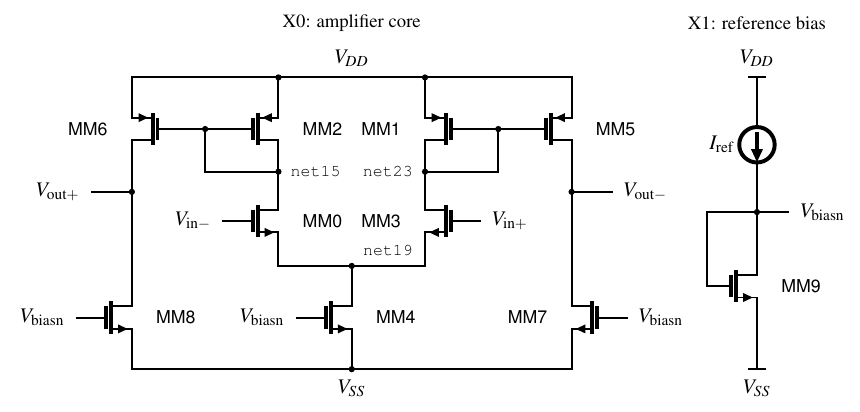}
\caption{Fully differential current-mirror amplifier used in Case~2.
X0 contains the amplifier and X1 supplies its shared bias voltage.
Device identifiers and connections follow the supplied netlist.
Identical bias labels denote the same electrical net.
In both schematics, $V_{\mathrm{in+}}$/$V_{\mathrm{in-}}$ and
$V_{\mathrm{out+}}$/$V_{\mathrm{out-}}$ correspond to
\texttt{Vinp}/\texttt{Vinn} and \texttt{Voutp}/\texttt{Voutn}, respectively.}
\label{fig:case2_schematic}
\end{figure}

\begin{taskbox}{Task Setup}
\begin{tabularx}{\linewidth}{@{}l>{\raggedright\arraybackslash}X@{}}
\toprule
\textbf{Topology} & Fully-differential current-mirror amplifier \\
\textbf{Evaluation split} & Out-of-distribution \\
\textbf{Technology} & 65~nm CMOS (PTM), $V_{DD}=1.2$~V \\
\midrule
\textbf{Gain target} & $\geq 26.4$~dB \\
\textbf{Power target} & $\leq 170.1~\mu$W \\
\textbf{Bandwidth target} & $\geq 1.25$~MHz ($-3$~dB bandwidth) \\
\textbf{Phase-margin target} & $30^\circ \leq \mathrm{PM} \leq 90^\circ$ \\
\textbf{Evaluation tolerance} & $5\%$, as defined in the main evaluation \\
\midrule
\textbf{Initial gain} & $-123.2$~dB \\
\textbf{Initial power} & $52~\mu$W \\
\textbf{Initial bandwidth} & Not measurable \\
\textbf{Initial phase margin} & Unavailable \\
\textbf{Initial in-band peaking} & $61.2$~dB \\
\textbf{Initial validity} & Invalid under the domain-validity gates. \\
\bottomrule
\end{tabularx}
\end{taskbox}

\begin{obsbox}{Structured System Input Excerpt}
\begin{verbatim}
<GRAPH_INFO>
TOPOLOGY_NAME: Fully_differential_current_mirror
</GRAPH_INFO>
<GRAPH_NODE_IDMAP>
type=pin: @0=vdd @1=vss
type=current_source: @55=X1.I0
type=n_mosfet: @22=X0.MM3 @27=X0.MM4 @42=X0.MM7 @47=X0.MM8 @58=X1.MM9 @7=X0.MM0
type=p_mosfet: @12=X0.MM1 @17=X0.MM2 @32=X0.MM5 @37=X0.MM6
type=net: @2=LG_Vbiasn @3=Vinn @4=Vinp @5=Voutn @6=Voutp ...
</GRAPH_NODE_IDMAP>
<PINMAP> ... </PINMAP>
<EDGES> ... </EDGES>

Design bounds:
For tech 65nm, length l must stay within [45nm, 100um] and width w must stay
within [45nm, 500um]. Designs outside this range are invalid.
Fixed global: vdd=1.2V (user-set, not tunable).

Allowed valid target handles and parameter tokens:
  @7: l_MM30, w_MM30      (MM0 -- NMOS input pair, matched with @22)
  @12: l_MM21, w_MM21     (MM1 -- PMOS diode load, matched with @17)
  @27: l_MM4, w_MM4       (MM4 -- NMOS tail)
  @32: l_MM56, w_MM56     (MM5 -- PMOS output mirror, matched with @37)
  @42: l_MM78, w_MM78     (MM7 -- NMOS output sink, matched with @47)
  @55: iref               (I0 -- bias current source)
  @58: l_MM9, w_MM9       (MM9 -- NMOS bias diode)
\end{verbatim}
\end{obsbox}

\begin{obsbox}{Turn 1 User Input Excerpt}
\begin{verbatim}
Your goal is: Achieve gain >= 26.4 dB, power <= 170.1 uW,
-3dB bandwidth >= 1.25 MHz,
30.0 deg <= phase_margin <= 90.0 deg
(each spec has 5% tolerance). Prioritize gain first.
[...]

Your current observation is:
<PERFORMANCE>
<GainAt1kHz>-123.2 dB</GainAt1kHz>
<Power>52 uW</Power>
<Bandwidth3dB>not measurable</Bandwidth3dB>
<InBandFlatness>61.2 dB (limit 6)</InBandFlatness>
<Validity>non-positive gain; power/bandwidth/PM not meaningful until gain >
  0</Validity>
</PERFORMANCE>
<FEATURES>
@7 id=X0.MM0 type=n_mosfet w=w_MM30=462.5n l=l_MM30=185.9n Vgs=647 mV
    Vds=970 uV Vdsat=132 mV Ids=188 nA Gm=1.3 uS region=triode
@12 id=X0.MM1 type=p_mosfet w=w_MM21=1.6u l=l_MM21=20u Vgs=-646 mV
    Vds=-646 mV Vdsat=-276 mV Ids=-188 nA Gm=1.2 uS region=sat
@27 id=X0.MM4 type=n_mosfet w=w_MM4=1.8u l=l_MM4=20u Vgs=522 mV
    Vds=553 mV Vdsat=127 mV Ids=428 nA Gm=5.5 uS region=sat
@32 id=X0.MM5 type=p_mosfet w=w_MM56=22u l=l_MM56=417.4n Vgs=-646 mV
    Vds=-32.0 uV Vdsat=-277 mV Ids=-30.5 nA Gm=102 nS region=triode
@42 id=X0.MM7 type=n_mosfet w=w_MM78=58.7n l=l_MM78=7.9u Vgs=522 mV
    Vds=1.2 V Vdsat=127 mV Ids=29.9 nA Gm=381 nS region=sat
@55 id=X1.I0 type=current_source dc=iref=8.2u
@58 id=X1.MM9 type=n_mosfet w=w_MM9=196.1n l=l_MM9=140.5n Vgs=522 mV
    Vds=522 mV Vdsat=118 mV Ids=8.1 uA Gm=93.6 uS region=sat
</FEATURES>
\end{verbatim}
\end{obsbox}
\begin{reasonbox}{Turn 1 Original LLM Output Excerpt}
\begin{verbatim}
<think>
The current operating point is invalid: DC gain is -123.2 dB (magnitude only
0.123 dB), power 52 uW is fine but bandwidth is not measurable because the
low-frequency gain is essentially zero. The main problems are:
1. PMOS load MM1/MM2 is very weak: W/L only 1.6u/20u = 0.08. Needs stronger PMOS
   current source.
2. NMOS pair MM3 is too weak: W/L only 462.5n/185.9n = 2.49, needs wider input
   pair.
3. Tail MM4 is poorly sized: W/L only 1.8u/20u = 0.09, should provide more tail
   current.
4. PMOS current source MM5/MM6 is too wide/triode: 22u/417n acts as resistor
   rather than DC current source.
5. NMOS load MM7/MM8 is too weak: W=58.7n, L=7.9u gives poor DC level shifting.
   Need stronger NMOS load.
6. iref can be reduced later, but first establish DC bias. Strategy: Make MM3
   strong pair, MM1/MM2 moderate PMOS load, stronger tail, better PMOS current
   source, stronger NMOS load, moderate iref.
</think>
<action>increase @22 w_MM30 500%</action>
<action>decrease @22 l_MM30 20%</action>
<action>increase @12 w_MM21 500%</action>
<action>decrease @12 l_MM21 99%</action>
<action>increase @27 w_MM4 500%</action>
<action>decrease @27 l_MM4 99%</action>
<action>decrease @32 w_MM56 95%</action>
<action>increase @32 l_MM56 100%</action>
<action>increase @42 w_MM78 500%</action>
<action>decrease @42 l_MM78 99%</action>
<action>decrease @55 iref 50%</action>
<action>increase @58 w_MM9 100%</action>
<action>increase @58 l_MM9 50%</action>
\end{verbatim}
\end{reasonbox}
\begin{resultbox}{Simulator Result After Turn 1}
\begin{tabular}{@{}ll@{}}
\textbf{Gain / power} & $-118.4$~dB / $25~\mu$W \\
\textbf{Bandwidth / phase margin} & Not measurable \\
\textbf{In-band peaking} & $39.5$~dB \\
\end{tabular}

\smallskip
The first update does not restore valid operation.
Gain remains non-positive and in-band peaking exceeds the
validity threshold. The simulator again returns a
\texttt{design\_error}.
\end{resultbox}
\begin{obsbox}{Turn 2 Observation Summary}
\begin{verbatim}
(observation: INVALID - design_error, gain -118.4 dB, power 25 uW,
 in-band flatness 39.5 dB; sizes after Turn 1: MM0/MM3 2.8u/148.7n, MM1/MM2
   9.6u/200n,
 MM4 10.8u/200n, MM5/MM6 1.1u/834.8n, MM7/MM8 352.2n/79n, MM9 392.2n/210.8n,
   iref 4.1u)
\end{verbatim}
\end{obsbox}
\begin{reasonbox}{Turn 2 Original LLM Output Excerpt}
\begin{verbatim}
<think>
The current operating point is invalid: DC gain is -118.4 dB (magnitude only
0.118 V/V). The DC gain polarity and magnitude are wrong. The main problems are
the bias: PMOS load MM1/MM2 is too strong compared to NMOS pair MM0/MM3, tail
MM4 current is moderate, PMOS current source MM5/MM6 and NMOS current mirror
MM7/MM8 are not properly sized. The DC operating point is such that Vout is
wrong. Need to restore DC: make NMOS pair MM0/MM3 stronger (larger W, longer L),
PMOS load MM1/MM2 weaker (smaller W, longer L), stronger tail MM4, stronger PMOS
current source MM5/MM6, and better NMOS current mirror MM7/MM8. iref can be
reduced later for power.
</think>
<action>increase @22 w_MM30 500.0%</action>
<action>increase @22 l_MM30 200.0%</action>
<action>decrease @12 w_MM21 80.0%</action>
<action>increase @12 l_MM21 400.0%</action>
<action>increase @27 w_MM4 200.0%</action>
<action>decrease @27 l_MM4 10.0%</action>
<action>increase @32 w_MM56 300.0%</action>
<action>decrease @32 l_MM56 50.0%</action>
<action>increase @42 w_MM78 500.0%</action>
<action>increase @42 l_MM78 100.0%</action>
<action>decrease @55 iref 20.0%</action>
<action>decrease @58 w_MM9 20.0%</action>
<action>decrease @58 l_MM9 20.0%</action>
\end{verbatim}
\end{reasonbox}
\begin{resultbox}{Simulator Result After Turn 2}
\begin{tabular}{@{}ll@{}}
\textbf{Gain / power} & $40.2$~dB / $224~\mu$W \\
\textbf{Bandwidth / phase margin} & $6.9$~MHz / $8.5^\circ$ \\
\textbf{In-band peaking} & $0.0$~dB \\
\end{tabular}

\smallskip
The design becomes valid and exceeds the gain and bandwidth
targets. Power remains above the budget, and phase margin is
below the accepted range.
\end{resultbox}
\begin{obsbox}{Turn 3 Observation Summary}
\begin{verbatim}
(observation: gain 40.2 dB, power 224 uW, bandwidth 7 MHz, PM 8 deg; MM0/MM3
  16.8u/446.1n
 Ids=4 uA, MM5/MM6 4.4u/417.4n Ids=-24.2 uA, MM7/MM8 2.1u/158n Ids=24.2 uA, iref
   3.3u)
\end{verbatim}
\end{obsbox}
\begin{reasonbox}{Turn 3 Original LLM Output Excerpt}
\begin{verbatim}
<think>
Current specs: Gain is 40.2 dB (well above 26.4 dB target), BW is 7 MHz (meets
1.25 MHz requirement), but Power is 224 uW (exceeds 170.1 uW) and Phase Margin
is only 8 deg (needs 30-90 deg). The main issues are excessive bias current and
over-sized PMOS current source/load devices causing too much unity-gain
frequency phase lag. Need to reduce iref, scale down MM5/MM6 PMOS current
source, MM7/MM8 NMOS load, and iref/MM9 bias to save power and improve phase
margin while maintaining gain and bandwidth.
</think>
<action>decrease @55 iref 65.0%</action>
<action>decrease @32 w_MM56 60.0%</action>
<action>decrease @32 l_MM56 40.0%</action>
<action>decrease @42 w_MM78 50.0%</action>
<action>decrease @42 l_MM78 20.0%</action>
<action>decrease @27 w_MM4 40.0%</action>
<action>decrease @27 l_MM4 15.0%</action>
<action>decrease @12 w_MM21 20.0%</action>
<action>decrease @12 l_MM21 20.0%</action>
<action>decrease @58 w_MM9 30.0%</action>
<action>increase @58 l_MM9 10.0%</action>
\end{verbatim}
\end{reasonbox}
\begin{resultbox}{Simulator Result After Turn 3}
\begin{tabular}{@{}ll@{}}
\textbf{Gain / power} & $36.2$~dB / $90~\mu$W \\
\textbf{Bandwidth / phase margin} & $4.8$~MHz / $23.6^\circ$ \\
\end{tabular}

\smallskip
The combined bias and sizing updates bring power within the
budget and improve phase margin from $8.5^\circ$ to $23.6^\circ$.
Only the phase-margin specification remains unmet.
\end{resultbox}
\begin{obsbox}{Turn 4 Observation Summary}
\begin{verbatim}
(observation: gain 36.2 dB, power 90 uW, bandwidth 5 MHz, PM 24 deg; MM0/MM3
  16.8u/446.1n
 Ids=2.1 uA, MM5/MM6 1.8u/250.4n Ids=-9.4 uA, MM7/MM8 1u/126.4n Ids=9.4 uA, iref
   1.2u)
\end{verbatim}
\end{obsbox}
\begin{reasonbox}{Turn 4 Original LLM Output Excerpt}
\begin{verbatim}
<think>
Current specs: Gain is 36.2 dB (well above 26.4 dB target), Power is only 90 uW
(well within 170.1 uW), Bandwidth is 5 MHz (exceeds 1.25 MHz requirement). The
only unmet spec is Phase Margin, which is 24 deg but requires 30.0 deg to 90.0
deg. The main issue is excessive DC gain due to large ro of signal path devices,
causing too much phase lag at unity-gain frequency. Need to reduce ro of
output/load devices to lower DC gain and improve phase margin while maintaining
gain and bandwidth. Strategy: reduce device lengths (MM0/MM3, MM1/MM2, MM5/MM6,
MM7/MM8) and iref to lower ro and DC current, decreasing DC gain and improving
phase margin. Decrease iref moderately to save power.
</think>
<action>decrease @7 w_MM30 50.0%</action>
<action>decrease @7 l_MM30 40.0%</action>
<action>decrease @12 w_MM21 30.0%</action>
<action>decrease @12 l_MM21 40.0%</action>
<action>decrease @32 w_MM56 50.0%</action>
<action>decrease @32 l_MM56 40.0%</action>
<action>decrease @42 w_MM78 40.0%</action>
<action>decrease @42 l_MM78 30.0%</action>
<action>decrease @55 iref 30.0%</action>
<action>decrease @27 w_MM4 30.0%</action>
<action>decrease @27 l_MM4 15.0%</action>
<action>decrease @58 w_MM9 20.0%</action>
<action>decrease @58 l_MM9 10.0%</action>
\end{verbatim}
\end{reasonbox}
\begin{resultbox}{Final Simulator Result After Turn 4}
\begin{tabular}{@{}ll@{}}
\textbf{Gain / power} & $28.4$~dB / $78.8~\mu$W \\
\textbf{Bandwidth / phase margin} & $10.2$~MHz / $46.8^\circ$ \\
\textbf{In-band peaking} & $0.0$~dB \\
\end{tabular}

\smallskip
All four original targets are satisfied without relying on
the evaluation tolerance. Gain remains $2.0$~dB above the target,
while phase margin reaches $46.8^\circ$ and power remains within
the budget. The episode terminates after four turns.
\end{resultbox}

These examples illustrate how simulator feedback can guide
recovery from invalid operation and subsequent refinement
of coupled performance metrics.
The two rollouts demonstrate successful optimization from different
initial conditions and specification targets on an in-distribution
topology and a held-out topology.
\par
\endgroup

\end{document}